\documentclass[journal]{IEEEtran}

\usepackage{graphicx}
\usepackage{subfigure}
\usepackage{lipsum}
\usepackage[noadjust]{cite}
\usepackage{verbatim}
\usepackage{amsmath}
\usepackage{setspace}   
\usepackage{fancyhdr}
\usepackage{color}
\usepackage{amsfonts}
\usepackage{tikz}
\usepackage{amsfonts}
\usepackage{url}
\usepackage{bm}
\usepackage[normalem]{ulem} 

\usepackage{booktabs}
\usepackage{tabularx}
\usepackage{ragged2e}
\usepackage{array} 
\newcolumntype{Y}{>{\RaggedRight\arraybackslash}X} 
\usepackage{makecell}
\usepackage{multirow}

\usepackage{threeparttable}

\usepackage[colorlinks=true,linkcolor=blue,citecolor=blue,urlcolor=blue]{hyperref}

\usepackage[ruled]{algorithm2e} 

\SetAlFnt{\small}
\SetAlCapFnt{\small}
\SetAlCapNameFnt{\small}
\SetAlCapHSkip{0pt}
\usepackage{stfloats}      
\usepackage{placeins}      
\usepackage{booktabs} 
\usepackage{textcomp}
\usepackage{wrapfig}
\usepackage{physics}
\usepackage{multirow}
\usepackage{xcolor}
\graphicspath{{./fig/}}
\DeclareGraphicsExtensions{.jpg}

\newcommand{\charlie}[1]{\textbf{\textcolor[rgb]{0.60,0.00,0.00}{[Charlie::~#1]}}}

\definecolor{revcolor}{rgb}{0.00,0.00,0.0}

\newcommand{\rev}[1]{\textcolor{revcolor}{#1}}

\newcommand{\Tian}[1]{\textbf{\textcolor[rgb]{0.412,0.682,0.782}{[Yingjun::~#1]}}}

\definecolor{mypurple}{RGB}{128,0,128}
\begin{document}
\title{Model-Free Co-Optimization of Manufacturable Sensor Layouts and Deformation Proprioception}


\author{
Yingjun Tian, Guoxin Fang,~\IEEEmembership{Member,~IEEE}, Aoran Lyu, Xilong Wang,  Zikang Shi, \\Yuhu Guo, Weiming Wang and Charlie C.L. Wang,~\IEEEmembership{Senior Member,~IEEE}   
\thanks{
Y. Tian, A. Lyu, X. Wang, Z. Shi, Y. Guo, W. Wang and C.C.L. Wang are all with the Department of Mechanical and Aerospace Engineering, The University of Manchester, United Kingdom (email: yingjun.tian@manchester.ac.uk; aoran.lyu@postgrad.manchester.ac.uk; \mbox{xilong.wang@postgrad.manchester.ac.uk}; guoyuhucv@gmail.com; zikang.shi@postgrad.manchester.ac.uk;  wwmdlut@gmail.com; 
charlie.wang@manchester.ac.uk).

G. Fang is with the Department of Mechanical and Automation Engineering, The Chinese University of Hong Kong, Shatin, Hong Kong. (email: guoxinfang@mae.cuhk.edu.hk).


\textit{Corresponding author}: Charlie C.L. Wang.
}
}

\markboth{Author's Version}
{\MakeLowercase{Tian \textit{et al.}}: Model-Free Co-Optimization of Manufacturable Sensor Layouts and Deformation Proprioception}

\maketitle

\begin{abstract}
Flexible sensors are increasingly employed in soft robotics and wearable devices to provide proprioception of freeform deformations. Although supervised learning can train shape predictors from sensor signals, prediction accuracy strongly depends on sensor layout, which is typically determined heuristically or through trial-and-error. This work introduces a model-free, data-driven computational pipeline that jointly optimizes the number, length, and placement of flexible length-measurement sensors together with the parameters of a shape prediction network for large freeform deformations. Unlike model-based approaches, the proposed method relies solely on datasets of deformed shapes, without requiring \rev{physical simulation models}, and is therefore broadly applicable to diverse robotic sensing tasks. The pipeline incorporates differentiable loss functions that account for both prediction accuracy and manufacturability constraints. By co-optimizing sensor layouts and network parameters, the method significantly improves deformation prediction accuracy over unoptimized layouts while ensuring practical feasibility. The effectiveness and generality of the approach are validated through numerical and physical experiments on multiple soft robotic and wearable systems.
\end{abstract}


\IEEEpeerreviewmaketitle
\section{Introduction}
Soft sensors can be designed and manufactured using conductive materials to withstand significant deformation, making them ideal for sensing tasks in applications involving large shape changes -- such as wearable devices \cite{Nag2017Wearable}, human–machine interfaces \cite{heng2022flexible}, and soft robotics \cite{Thomas2019Sensor,wang2024sensing}. These sensors are typically composed of elastomeric bodies such as silicone integrated with microfluidic channels filled with liquid metals like EGaIn and Galinstan \cite{dickey2017stretchable,park2012design}, by which the sensors can sustain strains of up to 200\%. Significant advances have been made in developing novel soft sensing materials \cite{yang2018hydrogel,kwon2019direct}, innovative structural and functional designs \cite{yamada2011stretchable,Yuki2016FunctionalSensor,yang2020laser}, and advanced manufacturing techniques \cite{bariya2018roll,liu20213d}. A comprehensive review of these developments can be found in \cite{luo2023technology}.




\subsection{Problem statement}\label{subsecProbDef}
\begin{figure*}
\centering
\includegraphics[width=1.01\textwidth]{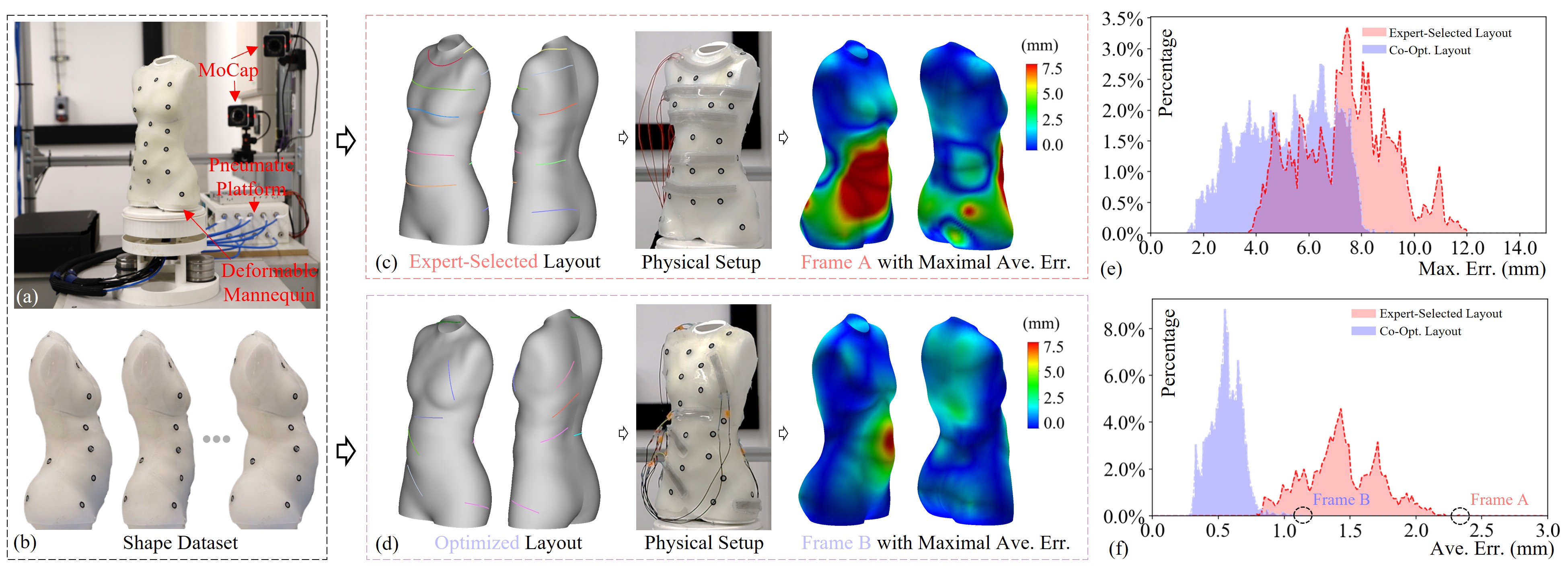}
\caption{For a pneumatic actuated deformable mannequin (a), the body shapes of different individuals, as shown in (b), can be realized and controlled with the help of stretchable sensors. When the sensor layout is determined by the intuitive design of experts in the garment industry -- i.e., heuristic consideration as the dimensions of girth measurement as shown in (c), large errors are observed in the shape predicted by these sensors (10 sensors with a total length of 1,204~mm). After applying the optimization approach presented in this paper, the error of shape prediction can be effectively reduced when using the same number of sensors but different layout as shown in (d) -- the total length reduced to 609~mm. The color map for shape approximation error on the models with maximal error are shown on the right of (c) and (d). The statistical comparisons of the shape prediction errors by using different sensor layouts can be found in (e) and (f) for the maximal errors and the average errors on 3,000 different shapes. 
}\label{figTeaser}
\end{figure*}

To effectively and efficiently reconstruct deformations from sensor signals, existing approaches often rely on machine learning models trained on datasets collected from physical prototypes (e.g.,~\cite{hu2023stretchable, Rob2021Sensing, shih2020electronic, Oliver2019DefCap, Oliver2019Golve}). Acquiring such datasets is generally time-consuming, involving sensor fabrication, calibration, and extensive experimental data collection. While recent quasi-static kinematics computation approaches (e.g., \cite{case2025expanded,Liu2025SoftRob,Fang2020TRO}) offer a more efficient alternative with the help of sim-to-real techniques (e.g., \cite{gao2024sim, Tian-RSS-24,Dubied2022RAL}), a more fundamental issue remains: the design of the sensor layout -- specifically, the number, shape, and placement of sensors -- is typically guided by heuristics or trial-and-error. As shown in the soft deformable mannequin example (Fig.~\ref{figTeaser}), heuristic sensor layouts produce significantly larger shape prediction errors than optimized layouts with the same number of sensors. 

The task of sensor-layout optimization is challenging due to the lack of a computational pipeline, as the accuracy of deformation prediction is coupled with both the mapping from sensing signals to deformed shapes and the arrangement of sensors. Only limited research has explored optimizing spatial sensor placement for accurately capturing freeform deformations with a minimal number of sensors. Prior works are constrained by different assumptions or formulations: simplified deformation models \cite{Kim2014CurvatureSensor}, user-specified number of sensors \cite{moritz2016DefSense}, selection-based approaches that cannot adjust sensor positions \cite{Javier2020MakeSense}, or voxel-based geometry representations \cite{Spielberg2021SensorOpt}, none of which support continuous optimization of sensor placement on curved surfaces. Furthermore, existing methods neglect practical manufacturability constraints -- challenges that are explicitly addressed in our approach.

This paper focuses on the co-optimization of layouts for highly stretchable resistive sensors, which are fabricated from liquid metal embedded in microfluidic channels and encapsulated in silicone (e.g., \cite{wang2024sensing}). While optimizing sensor layout is essential for improving prediction accuracy, the fabrication of such soft sensors imposes several practical design constraints.
\begin{itemize}
\item \textit{Length preference}: Each sensor must exceed a minimum length threshold $\bar{L}$ to function reliably. Very short sensors may fail to register deformation with sufficient sensitivity, whereas excessively long sensors increase manufacturing cost and complexity due to instabilities in liquid metal injection during fabrication\footnote{Straight sensors with length slightly greater than $\bar{L}=50\text{mm}$ are preferred in practice as they are easier to be fabricated.}. 

\item \textit{Inter-sensor distance}: The distance between any two sensors should exceed a threshold $\tau$, determined empirically for different materials. Insufficient spacing may introduce local stiffness discontinuities on curved surfaces, leading to stress concentrations and eventual material fatigue.

\item \textit{Overlap-free}: Sensor overlap must be avoided, as overlapping complicates fabrication and precise placement, and often introduces wrinkles on deformable surfaces that degrade sensing accuracy.
\end{itemize}
These constraints need be incorporated into the optimization process to ensure effective and manufacturable sensor layouts. Additionally, since the optimal shape prediction network depends on the specific sensor-layout, the layout and the shape predictor must be optimized jointly.

\subsection{Our method}
In this paper, we introduce a co-optimization framework that simultaneously computes the layout of sensors and the parameters of a shape prediction network, yielding an optimized number and placement of sensors that can accurately predict large freeform deformations while ensuring manufacturability. Without loss of generality, freeform deformable surfaces of soft robots or wearable devices are parameterized in the $u,v$-domain using a cubic B-spline surface representation with $u,v \in [0,1]$ by the method presented in \cite{Floater2000Meshless}. The shape of a deformable surface is encoded by its $m \times n$ control points $\mathcal{S}^c=\{\mathbf{P}^c_{i,j}\}$ forming a control mesh. For a dataset $\Theta$ with $N$ different shapes captured by their markers (see Fig.~\ref{figTeaser}(b) for an example), each model is represented by one such control mesh. Assuming up to $M$ sensors, each sensor is defined by its two endpoints in the $u,v$-domain. This representation of sample shapes and sensor locations is model-free and broadly applicable across robotic platforms, requiring no physical modeling or simulation in our co-optimization framework. \rev{Note that, although recent work on differentiable geodesic curve computation~\cite{Li2024DiffGeodesic} enables the incorporation of geodesic curves into gradient-based optimization, we choose to represent sensors as straight lines in the $(u,v)$-domain rather than as geodesic curves on the surface. This choice is motivated by the observation that geodesic curves may bypass salient surface features (such as `peaks' or `cavities' on freeform surfaces), which are the geometric details to be captured by sensors for shape prediction.} 

Giving the lengths measured by $M$ sensors on a deformed shape $\mathcal{S}$, denoted by a vector $\mathbf{s} \in \mathbb{R}^M$, we train a neural network $\mathcal{N}_{p}$ to predict the deformed shape directly from these measurements. Specifically, the predicted shape can be represented by the position of $m \times n$ control points, denoted as $\mathcal{S}^p = \mathcal{N}_{p}(\mathbf{s})$. The ground-truth shape of $\mathcal{S}$ is also compactly represented by its control points $\mathcal{S}^c$, allowing the shape prediction accuracy to be quantified as the discrepancy between $\mathcal{S}^p$ and $\mathcal{S}^c$. This accuracy is typically evaluated over a set of test shapes stored in the dataset $\Theta$. 

We propose a data-driven computational pipeline for optimizing deformable proprioception, which is predicted by the sensed length signals. Besides of the shape prediction errors, we also need to consider the manufacturing constraints in our formulation. Our co-optimization framework will be introduced in Sec.~\ref{secCompFramework} with details of loss functions presented in Sec.~\ref{secLossFunc}. Specifically, the technical contributions of our work are summarized as follows:
\begin{itemize}
\item A computational pipeline that jointly optimizes flexible sensor layouts and the parameters of a neural network for deformation proprioception from sensor signals;

\item A unified differentiable formulation that integrates the discrete optimization of sensor number with the continuous optimization of sensor locations;

\item Differentiable constraint encodings, in which practical fabrication requirements are reformulated from inherently discrete conditions into differentiable loss functions.
\end{itemize}
We validate the method through experiments on multiple soft robotic and wearable prototypes, demonstrating its effectiveness across diverse motions and tasks with complex freeform deformations.

\section{Related Work}

\subsection{Flexible and stretchable sensors}
Flexible and stretchable sensors have been widely adopted in a range of applications to support and monitor daily activities. Numerous types of such sensors have been developed, including electrochemical sensors \cite{Zhang2015SensorChemical, alwarappan2010enzyme}, resistance-based sensors \cite{yamada2011stretchable, Danna2020SimpleSensor}, and capacitance-based sensors \cite{cohen2012highly, Oliver2019Golve}. These devices are designed to capture diverse physiological and kinematic signals. For example, some models monitor heart rate \cite{patterson2009flexible} and breathing patterns \cite{jiang2010pvdf}, while others target human movement, such as ankle and elbow motion \cite{nag2016flexible}, hand gestures \cite{Oliver2019Golve}, and gait analysis \cite{tao2012gait}. Their success largely stems from their flexibility, stretchability, and ability to conform to highly curved geometries. More discussion of these technologies can be found in the survey papers \cite{Nag2017Wearable,shih2020electronic}. 

Nevertheless, the performance of such sensors strongly depends on their routing and placement -- factors that are often determined heuristically and rarely optimized under explicit manufacturability constraints. To address this gap, we investigate the optimization of sensor network layouts for deformation proprioception of freeform surfaces while explicitly incorporating manufacturability considerations.

\subsection{Deformation predicted by flexible sensors}
Leveraging their compliance and sensing capabilities, flexible sensors can capture the geometry of deformable objects in various forms, including single 3D points, 3D point clouds, 3D curves, and kinematic trees. For example, the 3D tip position of a 2-DoF serial soft actuator -- a single 3D point -- was measured using resistive sensors in \cite{Thomas2019Sensor}. The stretchable e-skin developed in \cite{hu2023stretchable} enabled reconstruction of soft manipulator shapes as 3D point clouds. For curve-like shape reconstruction, Li et al. \cite{Li2020TMechSensing} proposed parallel dual fiber Bragg grating (FBG) arrays. In addition, the soft sensing suit presented in \cite{mengucc2014wearable} demonstrated reconstruction of the human body’s kinematic tree without reliance on camera-based motion capture systems.

In more advanced robotic applications, particularly motion control of soft robots based on proprioception \cite{Tian2022SoRoMannequin,Meng2024SoftControl}, it is crucial to efficiently capture signals and reconstruct full 3D freeform surfaces. Prior works have demonstrated handcrafted capacitive arrays and photo-resistive sensors for tracking surface or membrane deformations \cite{Oliver2019DefCap,Rob2021Sensing}. However, methods for determining optimized sensor layouts have not been systematically explored. Compared with tip, curve, or skeletal reconstruction, surface-level shape reconstruction is far more sensitive to sensor placement and routing, as both local coverage and global continuity directly affect accuracy. This sensitivity -- combined with the lack of manufacturability considerations in most prior studies -- motivates our investigation of resistive sensor layout optimization for proprioception of deformable surfaces.

\subsection{Sensor design for proprioception}
Early sensor-layout design approaches were largely heuristic and rule-based rather than objective-driven. For example, sensors were arranged on predefined anatomical landmarks using human expertise as prior knowledge \cite{yamada2011stretchable, Oliver2019DefCap}, or guided by heuristic rules to direct optimization \cite{moritz2016DefSense}. In continuum manipulators, the Piecewise Constant Curvature (PCC) model has been widely adopted to prescribe sensor placements \cite{Kim2014CurvatureSensor}. These strategies are computationally lightweight and interpretable, but they constrain reconstruction to simple geometric assumptions \rev{-- i.e., are difficult to handle complicated deformation for freeform shapes as demonstrated in this paper}. 

Progress toward automatic design of sensor layout introduces a fundamental challenge: sensor number is inherently discrete, whereas the placements of sensors lie in continuous space. Handling discrete and continuous variables simultaneously within a unified framework is technically difficult. To avoid this coupling, prior approaches \cite{Javier2020MakeSense, kim2024optimal} reformulated the problem as a selection task, defining a dense set of candidate sensors and reducing optimization to choosing a subset. \rev{Wu et al.~\cite{wu2024placement} propose an optimization framework for placing flexible proximity sensors in human–robot collaboration scenarios. Their approach initializes a dense set of candidate geodesic paths on the surface, and employs Bayesian optimization to identify effective sensor configurations.} While these methods bypass the discrete-continuous challenge, it introduces new problems of the combinatorial search complexity and the sensitivity to the initial candidate sets. 

More recently, learning-driven co-optimization frameworks have been proposed, where sensor layouts and reconstruction models are optimized jointly in a differentiable manner \cite{Spielberg2021SensorOpt}. However, their approach remains limitations in two aspects: they rely on voxel-based representations, which restrict optimization to discrete volumetric elements, and they neglect manufacturability constraints, often resulting in impractical layouts. 

To address these limitations, we propose a co-optimization pipeline that jointly optimizes sensor layout, manufacturability, and shape prediction on freeform surfaces represented in the UV domain of B-spline surfaces.

\section{Computational Framework}\label{secCompFramework}

\subsection{Parameterization}\label{subsecPara}

\subsubsection{Freeform shapes}
\begin{figure}[t]
\centering
\includegraphics[width=0.45\textwidth]{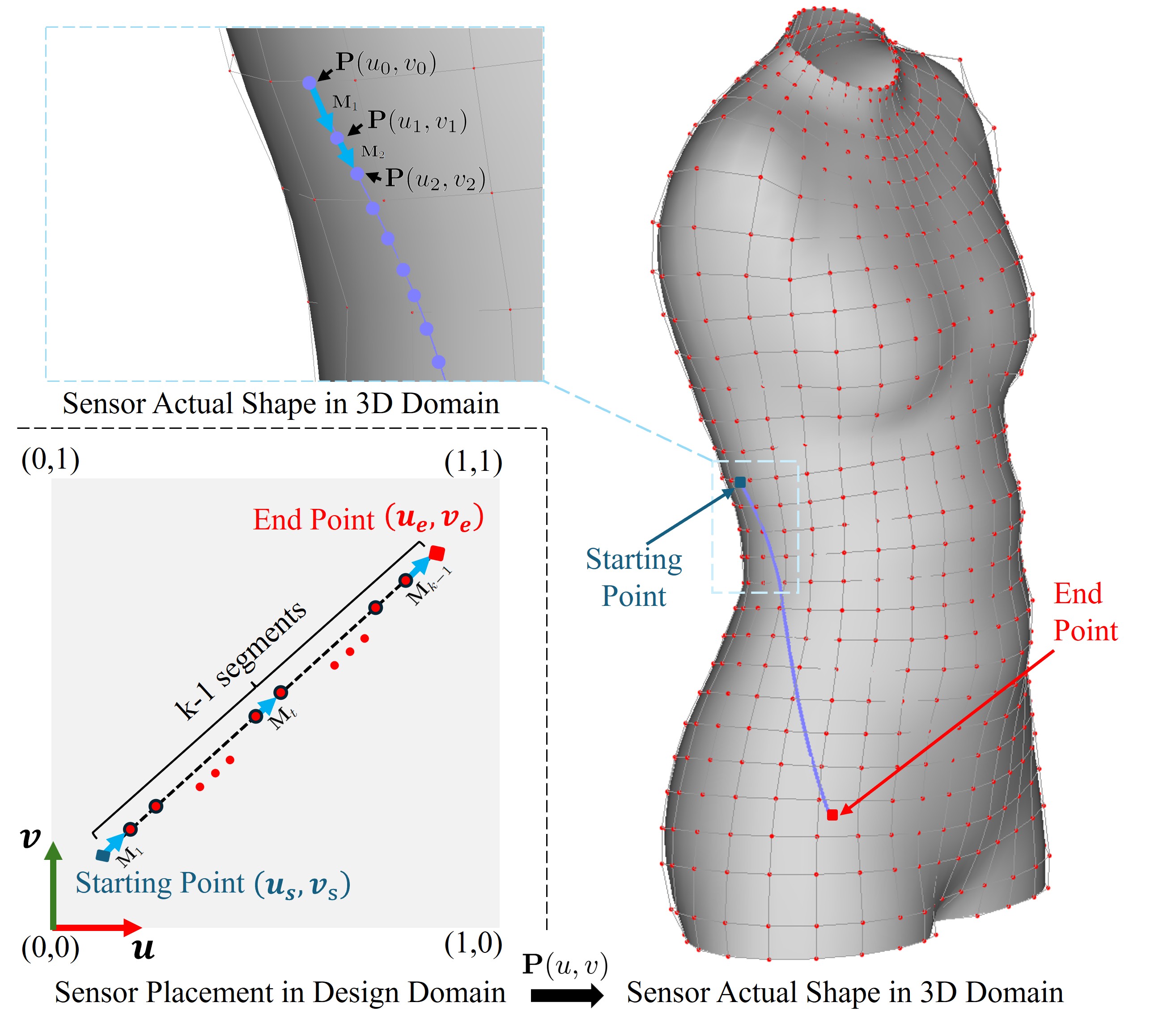}
\caption{The deformable freeform surface is parameterized by fitting a B-spline surface, defining a $u,v$-domain used as the design space for sensor layout optimization. Given the start point $(u_s,v_s)$ and the end point $(u_e,v_e)$, each sensor is represented as a line subdivided into $(K-1)$ segments, which are mapped onto the deformed 3D surface to compute its current length under deformation.
}\label{figDesignDomain}
\end{figure}

To support the computation of continuous gradient based optimization -- i.e., ensuring the differentiability of loss functions, we first conduct surface fitting to represent the shapes of a deformable robot as B-spline surfaces (ref.~\cite{Tian-RSS-24,mortenson1997geometric}) and then parameterize the location of sensors into the $u,v$-domain of B-spline surfaces. Specifically, each deformed shape $\mathcal{S}$ is represented as  
\begin{flalign}
\label{eqBSplineFunc}
\mathbf{S}(u,v) = \sum_{i=1}^{m}\sum_{j=1}^{n} N_{i}(u) N_{j}(v) \mathbf{P}^c_{i,j},
\end{flalign}
where a surface point $\mathbf{S}(u,v)$ is the linear sum of the B-spline basis functions ($N_{i}(u)$ and $N_{j}(v)$) and the control points $\{\mathbf{P}^c_{i,j}\}$. Similar to freeform surface representation in most computer-aided design systems, we choose cubic B-Spline in our work -- i.e., cubic polynomials are employed for $N_{i}(u)$ and $N_{j}(v)$. Each freeform surface can be compactly represented by 
$m \times n$ control points $\mathcal{S}^c=\{ \mathbf{P}^c_{i,j} \}$ as a polygonal mesh. An example of surface fitting result can be found in Fig.\ref{figDesignDomain}, where the B-spline surface is defined by a parametric domain with $u,v \in [0,1]$.

\subsubsection{Location and length-signal}
We represent each sensor as a straight line defined in the $u,v$-domain with two endpoints $(u_s,v_s)$ and $(u_e,v_e)$. When there are $M$ maximally allowed sensors, the layout of sensors are parameterized as a tensor $\mathbf{L}_{M \times 4}$ where the $k$-th sensor's location is stored in $\mathbf{L}$'s $k$-th row as $\mathbf{L}_k=[u_s \; v_s \; u_e \; v_e]$. For a length-measurement based flexible sensor, its corresponding signal -- i.e., the length of sensor can be effectively computed by subdividing the straight line into $K-1$ segments. Specifically, the $t$-th sample point will have its position in the $u,v$-domain as
\begin{equation}\label{eqLinePara}
u_t = \frac{u_e-u_s}{K-1}t+u_s, \; v_t = \frac{v_e-v_s}{K-1}t+v_s
\end{equation}
with $t=0,\ldots, K-1$. Therefore, the vector of the $t$-th line segment on the deformed surface is
\begin{flalign}
\label{eqLineSegment}
\mathbf{M}_t &=\mathbf{S}(u_t,v_t) - \mathbf{S}(u_{t-1},v_{t-1})  \nonumber \\ 
& = \sum_{i=1}^{m}\sum_{j=1}^{n} (N_{i}(u_{t}) N_{j}(v_{t}) - N_{i}(u_{t-1}) N_{j}(v_{t-1})) \mathbf{P}^c_{i,j}.
\end{flalign}
The sensor length on the deformed surface can be computed as the sum of every segment's length:
\begin{flalign}
\label{eqLineSegmentSum}
L_{s} = \sum_{t=1}^{K-1} \| \mathbf{M}_t  \| = \sum_{t=1}^{K-1} \sqrt{\mathbf{M}_{t}^T\mathbf{M}_{t}},
\end{flalign}
which is a function depending on $(u_s,v_s)$ and $(u_e,v_e)$ -- the variables to be optimized for sensor layout. The detail formulas for computing $\partial L_s / \partial u_s$, $\partial L_s / \partial v_s$, $\partial L_s / \partial u_e$ and $\partial L_s / \partial v_e$ can be found in Appendix A. 

\begin{figure*}
\centering
\includegraphics[width=1.0\textwidth]{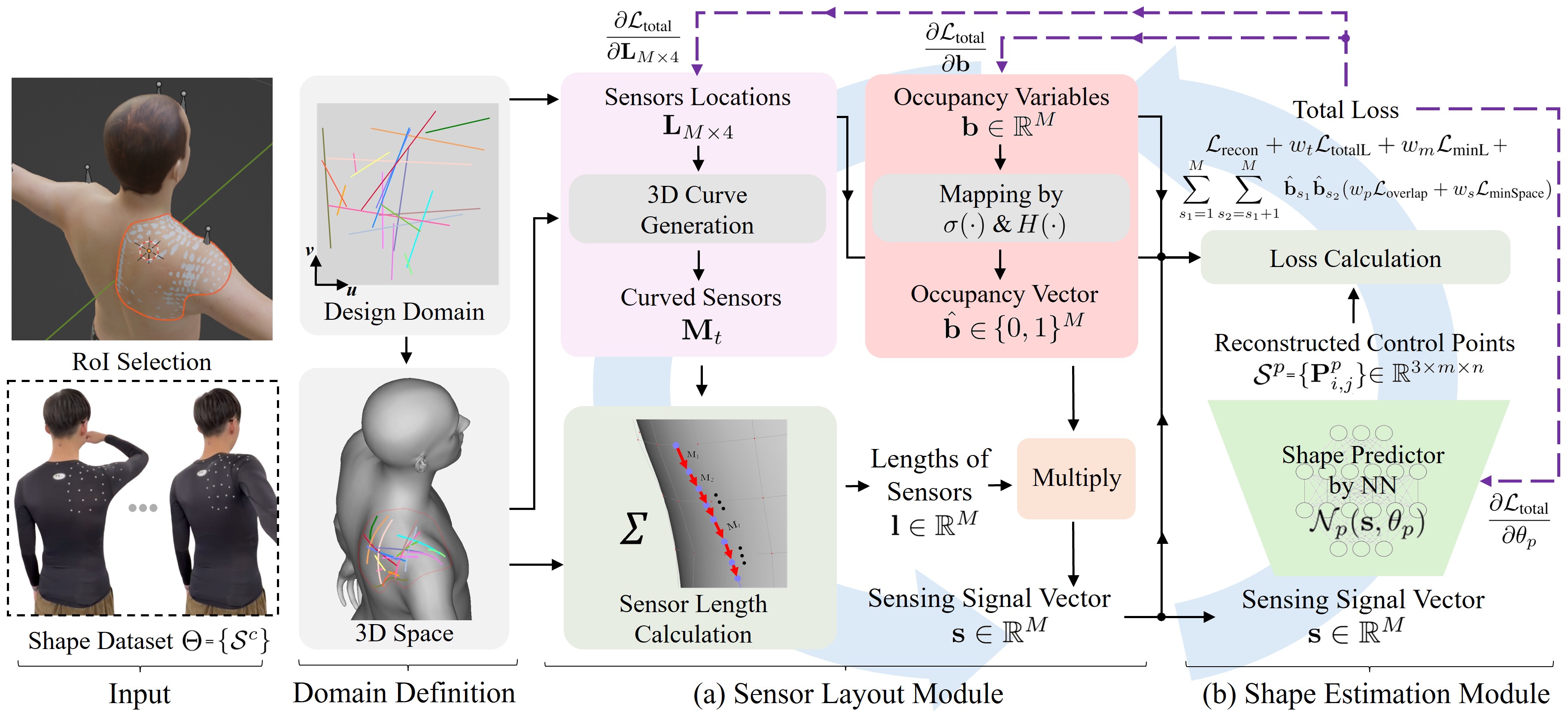}
\caption{Overview of our computational pipeline for co-optimizing (i) the sensor layout, parameterized by sensor locations $\mathbf{L}$ in the $u,v$-domain and the occupancy variable $\mathbf{b}$ \rev{(see (a) for the sensor layout module)}, and (ii) the shape prediction neural network $\mathcal{N}_p$, parameterized by its coefficients $\mathbf{\theta}_p$. Co-optimization is performed using a learning routine based on backpropagation over a training dataset of models with varying shapes $\Theta=\{\mathcal{S}^c\}$. The optimization is guided by loss functions that account for both shape prediction error and constraints related to design and manufacturability \rev{(see (b) for the shape estimation module)}. In the context of wearable applications, the input shapes are selected from regions-of-interest (RoI) on the human body.
}\label{figPipeline}
\end{figure*}

\subsubsection{Occupancy}
An occupancy vector $\hat{\mathbf{b}} \in \{0, 1\}^M$ is introduced to represent the \rev{existence} of each sensor\rev{, where a value of \textit{zero} indicates that the corresponding sensor can be removed}. To facilitate gradient-based optimization, we operate on a continuous variable $\mathbf{b} \in \mathbb{R}^M$. The values in $\mathbf{b}$ are first transformed into the range $[0, 1]$ using a sigmoid function, producing a relaxed occupancy vector. This representation is then projected to a binary vector mask in $\{0, 1\}$ using a Heaviside step function $H(\cdot)$, which is approximated by a smooth $\tanh$-based function as $H(x) \approx \frac{1}{2}(1+\tanh(\alpha x))$ to maintain differentiability during optimization. For the $k$-th sensor, we have
\begin{equation}
    \hat{\mathbf{b}}_k = \frac{1}{2}(1 + \tanh\left( \alpha( \sigma(\mathbf{b}_k) - 0.5)\right) ),
\end{equation}
with $\sigma(x) = 1 / (1 + e^{-x})$ being the sigmoid function and $\alpha$ being the parameter to tune the sharpness of Heaviside approximation. 

\subsection{Co-optimization Pipeline}\label{subsecCoOptmAlg}
After parameterizing the geometry of deformable freeform surface $\Theta = \{ \mathcal{S}^c \}$ and the sensor layouts $(\mathbf{L},\mathbf{b})$, we are able to conduct a computational pipeline to simultaneously optimize the sensor layouts $(\mathbf{L},\mathbf{b})$ and the neural network $\mathcal{N}_{p}$ for shape prediction. The method will be presented below with the help of illustration given in Fig.~\ref{figPipeline}.

Given the values of a parameterized sensor layout $\mathbf{L}$, we are able to compute the current lengths of all sensors as a vector $\mathbf{l} \in \mathbb{R}^M$, 
which is a function of $\mathbf{L}$ and $\mathcal{S}^c$. Together with $\mathbf{b}$, this will form the sensing signal vector $\mathbf{s} \in \mathbb{R}^M$ as input of the shape predictor network $\mathcal{N}_{p}$. The $k$-th component as the signal for the $k$-th sensor is weighted by the occupancy mask as
\begin{equation}
    \mathbf{s}_k = \hat{\mathbf{b}}_k \mathbf{l}_k. 
\end{equation}
With a well-trained neural network $\mathcal{N}_{p}$ parameterized on its coefficients $\mathbf{\theta}_p$, the shape of deformed surface can be predicted as
\begin{equation}\label{eqShapePredict}
    \mathcal{S}^p = \mathcal{N}_{p}(\mathbf{s}, \mathbf{\theta}_p).
\end{equation}
In short, the predicted shape $\mathcal{S}^p$ denoted by a set of control points $\{ \mathbf{P}^p_{i,j}\}$ $\in \mathbb{R}^{3\times m \times n}$ can be considered as a function of the layout parameters $(\mathbf{L},\mathbf{b})$ and the prediction parameter $\mathbf{\theta}_p$, which are variables to be optimized in our computational pipeline. 

The difference between $\mathcal{S}^p$ and $\mathcal{S}^c$ over all shapes in the training dataset $\mathcal{S}^c \in \Theta$ is used to supervise the co-optimization process via a loss function, combined with additional constraints described in Sec.~\ref{subsecProbDef}. The co-optimization problem is solved using a neural network-based computational pipeline, which benefits from automatic differentiation and stochastic gradient-based solvers. In other words, as long as the loss functions are differentiable with respect to the optimization variables -- namely $\mathbf{L}$, $\mathbf{b}$, and $\mathbf{\theta}_p$ -- the backpropagation mechanism of modern neural network training frameworks can be leveraged to solve the co-optimization problem. Details of the loss functions are presented in Sec.~\ref{secLossFunc}.

\section{Loss Functions}\label{secLossFunc}
A set of loss functions are defined in our computational pipeline to jointly optimize multiple objectives, including shape prediction accuracy, overlap avoidance, inter-sensor spacing, and length control.

\subsection{Shape prediction}
The shape prediction accuracy is optimized by introducing a loss function based on the Mean Squared Error (MSE) between the predicted control points and the ground-truth control points stored in the training dataset $\Theta$. Specifically, the reconstruction loss is defined as:
\begin{flalign}\label{eqLossRecon}
\mathcal{L}_{\text{recon}} := \frac{1}{mn|\Theta |} \sum_{\mathcal{S}^c \in \Theta} \sum_{i=1}^{m}\sum_{j=1}^{n} \| \mathbf{P}^p_{i,j} - \mathbf{P}^c_{i,j} \|^2,
\end{flalign}
where $\mathbf{P}^p_{i,j}$ and $\mathbf{P}^c_{i,j}$ denote the control points of the predicted shape $\mathcal{S}^p$ and the corresponding ground-truth shape $\mathcal{S}^c$, respectively. Here, $| \cdot |$ represents the cardinality of a set. According to the theoretical analysis of surface distance error between B-spline surfaces (ref.~\cite{Wang2014DynProgSurFit}), the MSE between control points gives a tight error-bound of the surface–surface Hausdorff distance. This justifies the use of control point-wise MSE as a reliable surrogate for surface similarity. Note that $\{ \mathbf{P}^p_{i,j}\}$ in $\mathcal{L}_{\text{recon}}$ is a function of the network coefficients $\mathbf{\theta}_p$ and the measured length-signals $\mathbf{s}$ (see Eq.\eqref{eqShapePredict}), where the sensor lengths $\mathbf{s}$ on each given model (i.e., $\{ \mathbf{P}^c_{i,j}\}$) is a function of layout parameters $(\mathbf{L},\mathbf{b})$.

\begin{figure}[t]
\centering
\includegraphics[width=0.99\linewidth]{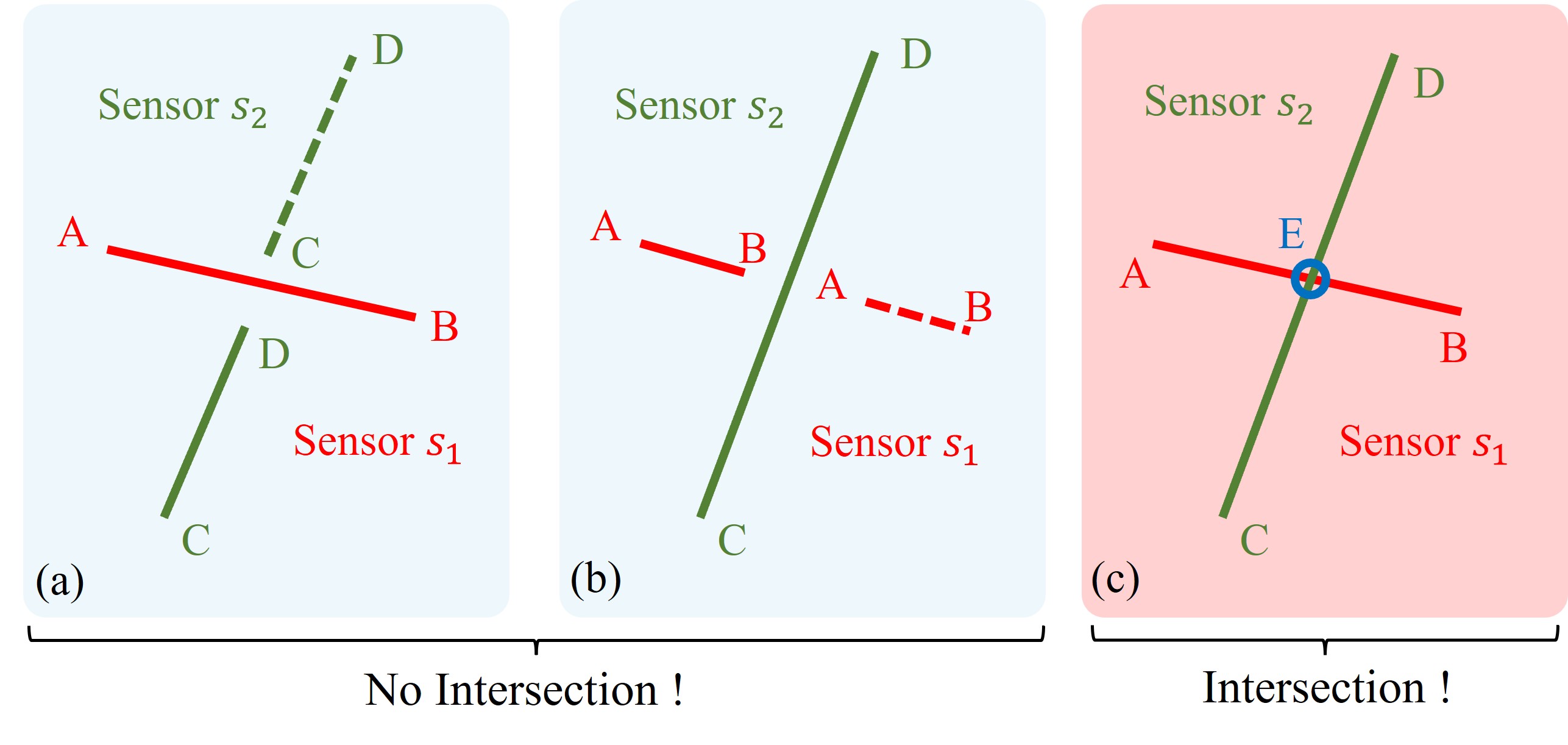}
\vspace{-10pt}
\caption{Possible intersection scenarios between Sensor $s_1$  (AB) and Sensor $s_2$ (CD), represented as straight line segments, include: (a, b) one sensor lies entirely on the same side of the other -- no intersection occurs; and (c) the endpoints of each sensor lie on opposite sides of the other, indicating a potential intersection.
}\label{figIntersection}
\end{figure}

\subsection{Overlap avoidance}
Since intersection between two sensors on a 3D freeform surface necessarily implies intersection in the $u,v$-domain, the overlap avoidance loss is formulated in the $u,v$-domain. As illustrated in Fig.~\ref{figIntersection}, for line segments defined in the $u,v$-domain, a necessary condition for the intersection between two line segments $AB$ and $CD$ is that each segment's endpoints must lie on opposite sides of the other \cite{cormen2022introduction}. This leads to the requirement that both of the following conditions must be satisfied:
\begin{flalign}\label{eqCrossingCondition}
\begin{cases} 
 (\overrightarrow{AD} \times \overrightarrow{CD})(\overrightarrow{BD} \times \overrightarrow{CD}) \leq 0, \\
(\overrightarrow{CB} \times \overrightarrow{AB})(\overrightarrow{DB} \times \overrightarrow{AB}) \leq 0. 
\end{cases} 
\end{flalign}
The first condition ensures that the endpoints of Sensor $s_1$  ($AB$) lie on opposite sides of Sensor $s_2$ ($CD$), while the second condition ensures the reverse -- that the endpoints of Sensor $CD$ lie on opposite sides of Sensor $AB$. Note that if any of the expressions in the conditions equal zero, it indicates a special case where an endpoint of one sensor lies exactly on the line segment of the other.

Given a Heaviside step function $H(\cdot)$, the loss term for overlap avoidance between sensor $s_1$ and $s_2$ can then be defined according to the above analysis as
\begin{flalign}\label{eqLossCross}
\mathcal{L}_\text{overlap} :&= H(f_1) H(f_2) \\
&  \approx \frac{1}{4}(1+\tanh (\alpha f_1))(1 +\tanh (\alpha f_2)) \nonumber
\end{flalign}
where
\begin{flalign}\label{eqF1}
f_1  = & \left(\left(u_{e1} - u_{e2}\right) \left(v_{e2} - v_{s2}\right) - \left(u_{e2} - u_{s2}\right) \left(v_{e1} - v_{e2}\right)\right) \nonumber \\ 
    & \left(\left(u_{e2} - u_{s1}\right) \left(v_{e2} - v_{s2}\right) - \left(u_{e2} - u_{s2}\right) \left(v_{e2} - v_{s1}\right)\right) \nonumber, \\
f_2 = & \left(\left(u_{e1} - u_{e2}\right) \left(v_{e1} - v_{s1}\right) - \left(u_{e1} - u_{s1}\right) \left(v_{e1} - v_{e2}\right)\right) \nonumber \\ 
& \left(\left(u_{e1} - u_{s1}\right) \left(v_{e1} - v_{s2}\right) - \left(u_{e1} - u_{s2}\right) \left(v_{e1} - v_{s1}\right)\right), \nonumber 
\end{flalign}
and $A=(u_{s1}, v_{s1})$, $B=(u_{e1}, v_{e1})$, $C=(u_{s2}, v_{s2})$ and $D=(u_{e2}, v_{e2})$ are the $u,v$-coordinates of the sensor $s_1$'s ending points and the sensor $s_2$'s ending points as the layout of sensors defined in $\mathbf{L}$. When either $f_1$ or $f_2 < 0$, it indicates that the corresponding endpoints lie on the same side and thus the loss $\mathcal{L}_\text{overlap}$ returns \textit{zero} -- i.e., meaning no intersection occurs. The function $\tanh(\cdot)$ is used to approximate the non-differentiable Heaviside step function, thereby enabling gradient-based optimization by ensuring smoothness and differentiability of the loss. This formulation enables efficient gradient computation of $\mathcal{L}_\text{overlap}$ w.r.t. sensor locations using symbolic differentiation (ref.~\cite{symbolicDiff}).

\begin{figure}[t]
\centering
\includegraphics[width=\linewidth]{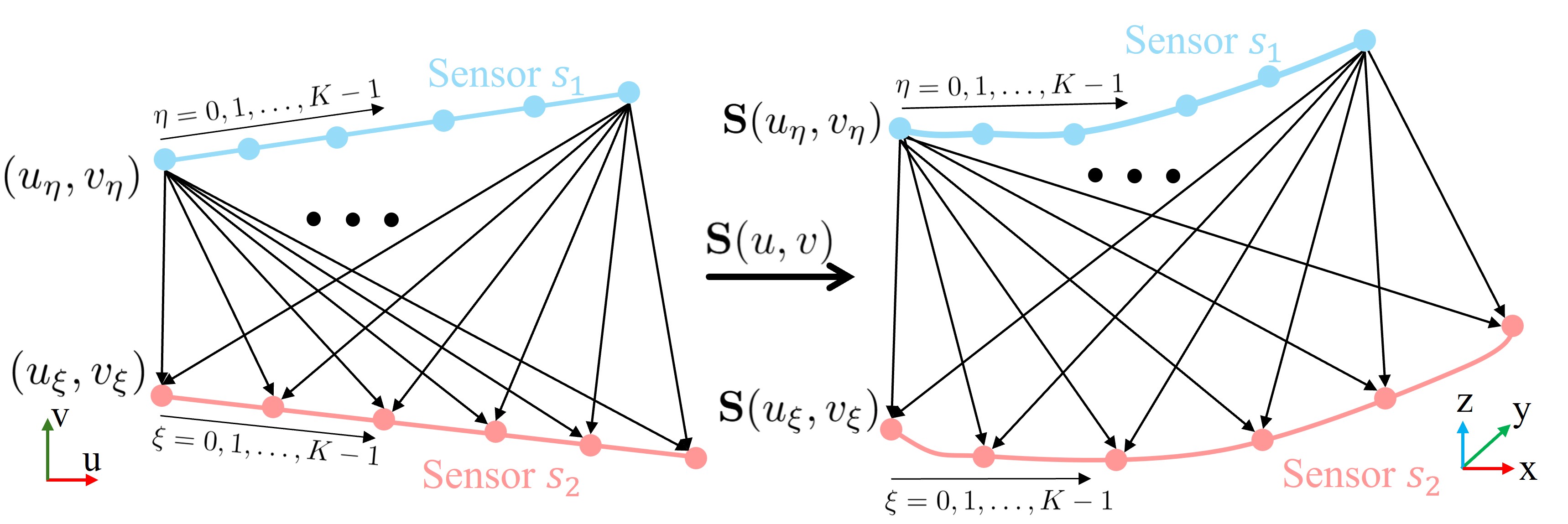}
\caption{
Illustration of computing the minimum distance between two sensors: (a) each sensor is first uniformly subdivided into $K$ points, denoted as $\{(u_\eta,v_\eta)\}$ and $\{(u_\xi,v_\xi)\}$, with 
$\eta$ and $\xi$ as the respective indices, and (b) the minimum Euclidean distance between the two sensors is obtained by exhaustively computing the pairwise distances between all points from the two sets mapped onto the surface. 
}\label{figMinimalDis}
\end{figure}

\subsection{Inter-sensor distance}
Sensor optimization based solely on $\mathcal{L}_\text{overlap}$ does not guarantee sufficient spacing between sensors. To address this, we introduce a minimum spacing loss, $\mathcal{L}_\text{minSpace}$, which quadratically penalizes configurations where the minimum distance $D(s_1,s_2)$ between two sensors $s_1$ and $s_2$ in 3D space (instead of $u,v$-domain) falls below a specified threshold $\tau$. 
\begin{equation}\label{eqMinimalDist}
\mathcal{L}_\text{minSpace} := (\tau - D(s_1,s_2))^2 H(\tau - D(s_1,s_2))
\end{equation}
Although the Heaviside function $H(\cdot)$ is not differentiable at the origin, the overall loss function remains $C^2$-continuous due to the squared difference term, which smooths the transition near the threshold. In our work, the value of $\tau = 10\text{mm}$ is determined by experimental tests of fabrication.

The distance $D(s_1,s_2)$ between two sensors $s_1$ and $s_2$ is computed with the help of sample points. Similar to the sampling method we introduced in Sec.~\ref{subsecPara}, the line segment for each sensor is sampled into $K-1$ segments with the $u,v$-coordinate of a sample can be computed by Eq.~\eqref{eqLinePara} -- see also Fig.~\ref{figMinimalDis}. Without loss of generality, we denote these samples as $(u_\eta,v_\eta)$ for points on the sensor $s_1$ and $(u_\xi,v_\xi)$ for points on the sensor $s_2$ with $\eta,\xi = 0,1, \ldots, K-1$. Minimum distance between the 3D curves of $s_1$ and $s_2$ can be computed by 
\begin{flalign}\label{eqSurfPntDist}
D(s_1,s_2) = \min_{\eta, \xi} \| \mathbf{S}(u_\eta,v_\eta) -  \mathbf{S}(u_\xi,v_\xi) \| 
\end{flalign}
with $\mathbf{S}(\cdot,\cdot)$ being the B-spline surface defined in Eq.\eqref{eqBSplineFunc}.
To make it differentiable, a soft minimum approximation is employed to define $D(s_1,s_2)$ as

\begin{equation}\label{eqSurfPntDistApprox}
   D \approx - \frac{1}{\beta} \log \bigg(  \sum_{\eta} \sum_{\xi} \exp \left( 
     -\beta \, \big\| \mathbf{S}(u_\eta,v_\eta) - \mathbf{S}(u_\xi,v_\xi) \big\| 
   \right) \bigg). 
\end{equation}
As such, $D(s_1,s_2)$ and also $\mathcal{L}_\text{minSpace}$ are differentiable in terms of the sensor's ending points -- i.e., $(u_{s1}, v_{s1})$, $(u_{e1}, v_{e1})$, $(u_{s2}, v_{s2})$ and $(u_{e2}, v_{e2})$ defined in the layout $\mathbf{L}$.

\subsection{Length control}
According to the design and manufacturing requirements, both the minimal length of each sensor and the total length of all sensors need to be controlled via optimization. The total length of all sensors is optimized by the loss function as:
\begin{flalign}\label{eqLossTotalL}
\mathcal{L}_{\text{totalL}} := \sum_{k=1}^M \hat{\mathbf{b}}_k \mathbf{l}_k
\end{flalign}
where the $k$-th sensor's length $\mathbf{l}_k$ on a surface $\mathbf{S}(u,v)$ can be computed by Eqs.~\eqref{eqLineSegment} and \eqref{eqLineSegmentSum}, and $\hat{\mathbf{b}}_k$ indicate the on / off mask of the $k$-th sensor. 

Given an allowed minimum length $\bar{L}_{\min}$, the segment length control can be enforced by the following loss term:
\begin{flalign}\label{eqLossMinL}
\mathcal{L}_{\text{minL}} :&= \sum_{k=1}^M \hat{\mathbf{b}}_k (\bar{L}_{\min} - \mathbf{l}_k)^2 H(\bar{L}_{\min} - \mathbf{l}_k).
\end{flalign}
where $\mathbf{l}_k$ denotes the length of the $k$-th sensor. This loss penalizes segments shorter than 
$\bar{L}_{\min}$. Notably, the overall function remains $C^2$-continuous near the threshold 
$\bar{L}_{\min}$, due to the quadratic smoothing term, and therefore does not require an approximation of the Heaviside function $H(\cdot)$.


\subsection{Total loss}
Combining above loss terms, we define the total loss as 
\begin{flalign}\label{eqTotal}
\mathcal{L}_{\text{total}} = & \mathcal{L}_{\text{recon}} + w_{t} \mathcal{L}_{\text{totalL}} + w_{m}\mathcal{L}_{\text{minL}}  \\
 & + \sum_{s_1, s_2 \in \Gamma} \hat{\mathbf{b}}_{s_1} \hat{\mathbf{b}}_{s_2}(w_{p} \mathcal{L}_\text{overlap} + w_{s} \mathcal{L}_\text{minSpace}) \nonumber
\end{flalign}
where $\Gamma$ denotes the collection of sensors. This total loss is employed in our computational pipeline (as illustrated in Fig.~\ref{figPipeline}) to supervise the optimization of the sensor layout $(\mathbf{L},\mathbf{b})$ and the neural network coefficients $\mathbf{\theta}_p$ via backpropagation of training. The weights $w_{t}=w_{s}=0.005$, $w_{m}=0.1$ and $w_{p}=0.6$ are determined by experimental tuning and employed to balance the weights between different loss terms. Details of training and weight tuning can be found in Sec.~\ref{subsecWeightTuning}. \rev{Note that we do not specifically control the number of sensors -- i.e., they are automatically changed via the occupancy variables $\{\hat{\mathbf{b}}_k\}$. When $\mathcal{L}_{\text{totalL}}$ is introduced, the optimizer tends to reduce the number of sensors indirectly while optimizing the surface prediction error via the loss $\mathcal{L}_\text{recon}$.}

\subsection{Computational Complexity}

\begin{table*}[t]
\centering
\footnotesize
\color{revcolor} 
\begin{threeparttable}
\caption{{Complexity analysis of the proposed co-optimization pipeline.$\dagger$}}
\label{tab:time_complexity}
\setlength{\tabcolsep}{6pt}
\renewcommand{\arraystretch}{1.25}

\begin{tabularx}{\textwidth}{@{} l|Y|l|Y @{}}

\toprule
\textbf{Component} & \textbf{What is computed} & \textbf{Time Complexity} & \textbf{Notes / Dependencies} \\
\midrule
{\textbf{Sensor length}} &
{Compute sensor length using $K$ sample points on each sensor (Eqs.~\eqref{eqLinePara}-\eqref{eqLineSegmentSum}).} &
{$\mathcal{O}(K)$} &
{$\mathcal{O}(1)$ complexity for each point due to B-spline's local support formulation.} \\

{\textbf{Reconstruction loss $\mathcal{L}_{\text{recon}}$}} &
{MSE evaluation over $m \times n$ control points of B-spline surface (Eq.~\eqref{eqLossRecon}).} &
{$\mathcal{O}(m n)$} &
{Linear to the total number of control points.} \\

{\textbf{Overlap avoidance $\mathcal{L}_\text{overlap}$}} &
{Pairwise intersection check among active sensors (Eqs.~\eqref{eqCrossingCondition}-\eqref{eqLossCross}).} &
{$\mathcal{O} (M^2)$} &
{Pairwise checks in the $(u,v)$ domain for $M_a$ active sensors.} \\

{\textbf{Inter-sensor distance $\mathcal{L}_\text{minSpace}$}} &
{Soft-min distance between sensor pairs by sample points (Eqs.~\eqref{eqMinimalDist}-\eqref{eqSurfPntDistApprox}).} &
{$\mathcal{O} (M^2 K^2)$} &
{For each sensor pair, compute distances between $K^2$ pairs of samples.} \\

{\textbf{Length control $\mathcal{L}_{\text{minL}}$, $\mathcal{L}_{\text{totalL}}$}} &
{Total length regularization and minimal length penalty (Eq.~\eqref{eqLossTotalL}-\eqref{eqLossMinL}).} &
{$\mathcal{O} (MK)$} &
{Similar to the complexity for evaluating the lengths of $K$ sensors.} \\

{\textbf{Shape prediction by $\mathcal{N}_p(\cdot)$}} &
{Forward computation of NN that maps sensor signals to the freeform shape as B-spline surface's control points.} &
{$\mathcal{O}((M + mn) H + L H^2)$} &
{$\mathcal{O}(MH)$ for the input layer, $\mathcal{O}(LH^2)$ for the hidden layers, and $\mathcal{O}(mnH)$ for the output layer predicting $m \times n$ control pnts.} \\
\bottomrule
\end{tabularx}
\vspace{2pt}
\begin{tablenotes}[leftmargin=1.5em]
\footnotesize
\item[$\dagger$] {$m \times n$ gives the number of control points on a B-spline surface;}

\item {$K$ presents the sampling density per sensor -- i.e., the number of points, and $M$ is the maximally allowed number of sensors;}

\item {$L$ and $H$ denote the number of hidden layers and the number of neurons per layer, respectively.} 
\end{tablenotes}

\end{threeparttable}
\end{table*}

\rev{The optimization of the total loss can be efficiently carried out using modern neural network optimizers (e.g., Adam~\cite{kingma2014adam}). We analyze below the computational complexity of evaluating each component in $\mathcal{L}_{\text{total}}$. Suppose that at most $M$ sensors are used, and each sensor is discretized into $K$ sample points. The length computation for a single sensor has complexity $\mathcal{O}(K)$, and thus the total length evaluation over all sensors has complexity $\mathcal{O}(MK)$. This efficiency stems from the fact that evaluating a 3D point on a B-spline surface with an $m \times n$ control net has constant complexity $\mathcal{O}(1)$, rather than $\mathcal{O}(mn)$, due to the compact local support of B-spline basis functions.}

\rev{The most computationally expensive term is the inter-sensor distance loss $\mathcal{L}_{\text{minSpace}}$, which requires evaluating distances between $\mathcal{O}(M^2)$ pairs of sensors. For each sensor pair, the distance computation between their sampled points has complexity $\mathcal{O}(K^2)$. Therefore, the overall computational cost grows quadratically w.r.t. both the number of samples per sensor $K$ and the maximum number of sensors $M$.}

\rev{The computational complexity of the neural network $\mathcal{N}_p(\cdot)$ for shape prediction is determined by the number of layers $L$ and the number of neurons $H$ per layer. Specifically, the input layer has complexity $\mathcal{O}(MH)$, the $L$ hidden layers contribute $\mathcal{O}(LH^2)$, and the output layer has complexity $\mathcal{O}(mnH)$. Overall, a single forward pass for shape prediction has complexity $\mathcal{O}((M + mn)H + LH^2)$. When using modern optimizers such as Adam~\cite{kingma2014adam} to update the parameters of $\mathcal{N}_p$ via backpropagation, the computational cost of automatic differentiation and parameter updates is of the same order as that of the forward pass (ref.~\cite{baydin2018automatic}). A summary of the resulting time complexity is provided in Table~\ref{tab:time_complexity}.}

\section{Implementation Details}\label{secImplementation}
In this section, we present the implementation details encompassing both software and hardware components of our system. The co-optimization of the sensor layout and the deformation prediction neural network is implemented in Python (version 3.8.17) using PyTorch (version 2.1.1), with a differentiable B-spline library integrated via Pybind. Once the co-optimization is completed, the optimized sensor layout and the trained neural network are exported to a C++ based shape prediction module, which is used during the inference phase for experimental validation. All training and computational evaluations are performed on a desktop computer equipped with an Intel i7-12700H CPU, an NVIDIA RTX 3060 GPU, and 32 GB of RAM.
The project page with the supplementary video and the source code of our implementation can be accessed at: \url{https://github.com/YingGwan/SensorOpt}.

\begin{figure}[t]
\centering
\includegraphics[width=1.0\linewidth]{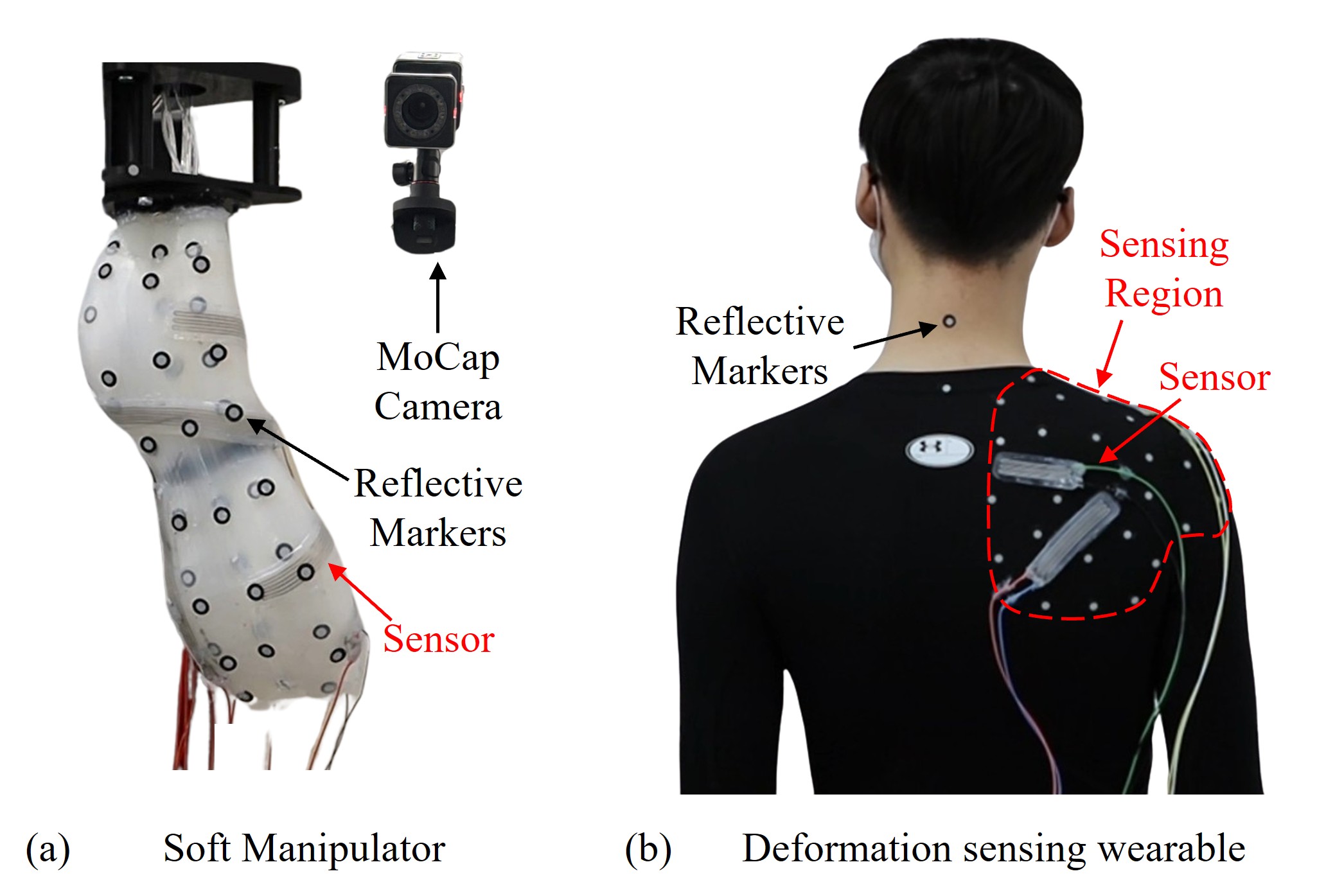}
\vspace{-15pt}
\caption{Two hardware setups used to validate the effectiveness of our co-optimization pipeline: (a) a soft manipulator consisting of two segments with a total of six pneumatic actuation chambers, and (b) a wearable device positioned over the shoulder region for sensing shoulder motion. 
}\label{figPhysicalSetups}
\end{figure}

\subsection{Experimental hardware} 
Three hardware setups are employed in our experimental evaluation: a soft deformable mannequin (Fig.~\ref{figTeaser}(a)), a soft manipulator (Fig.~\ref{figPhysicalSetups}(a)), and a shoulder sensing wearable (Fig.~\ref{figPhysicalSetups}(b)). In the first two soft robotic systems, pneumatic actuation is employed by using an open-source pneumatic control platform, 
which provides stable pressure output in the range of [-50.0kPa, 80.0kPa]. To capture ground-truth deformation data across all three setups, we employ a Vicon motion capture system equipped with 8 cameras. The 3D positions of reflective markers attached to each structure serve as reference data for evaluating deformation and validating shape prediction accuracy.

\subsubsection{Deformable mannequin}
The soft deformable robot used in this study was designed for customized garment fabrication \cite{Tian2022SoRoMannequin}. It is equipped with nine independent actuators distributed across the front and back surfaces, allowing deformation to match with the scanned body shape of individual clients. Given the complexity of a mannequin's 3D geometry, achieving accurate shape sensing with minimal number of sensors while adhering to fabrication constraints is challenging.

\subsubsection{Soft manipulator} 
The second hardware platform is a soft manipulator composed of two serially connected segments (ref.~\cite{Marchese2015ICRA}), each consisting of three pneumatic chambers, with a total length of 26.0 cm. The manipulator undergoes bending deformations that result in substantial changes to its outer surface. Traditional sensing approaches often simplify deformation modeling by capturing only the global pose, neglecting the rich surface deformations that occur during actuation. This simplification limits the ability of sensor arrays to accurately reconstruct freeform geometries under large deformations, which is critical for applications such as motion planning \cite{marchese2016design}. 

\subsubsection{Deformation sensing wearable}
The third platform is a soft, sensorized wearable designed for capturing complex human body motions and deformations. The wearable interface is fabricated by casting Ecoflex 0030 silicone 
onto a mold derived from a custom shoulder scan acquired using an Artec 3D scanner. The shoulder exhibits highly complex deformations due to the underlying muscle and bone structures, making it a  challenging region for accurate shape sensing. This setup provides a compelling demonstration of the proposed sensor optimization pipeline's capability to handle freeform, anatomically complex deformations.

\begin{figure}
\centering
\includegraphics[width=0.5\textwidth]{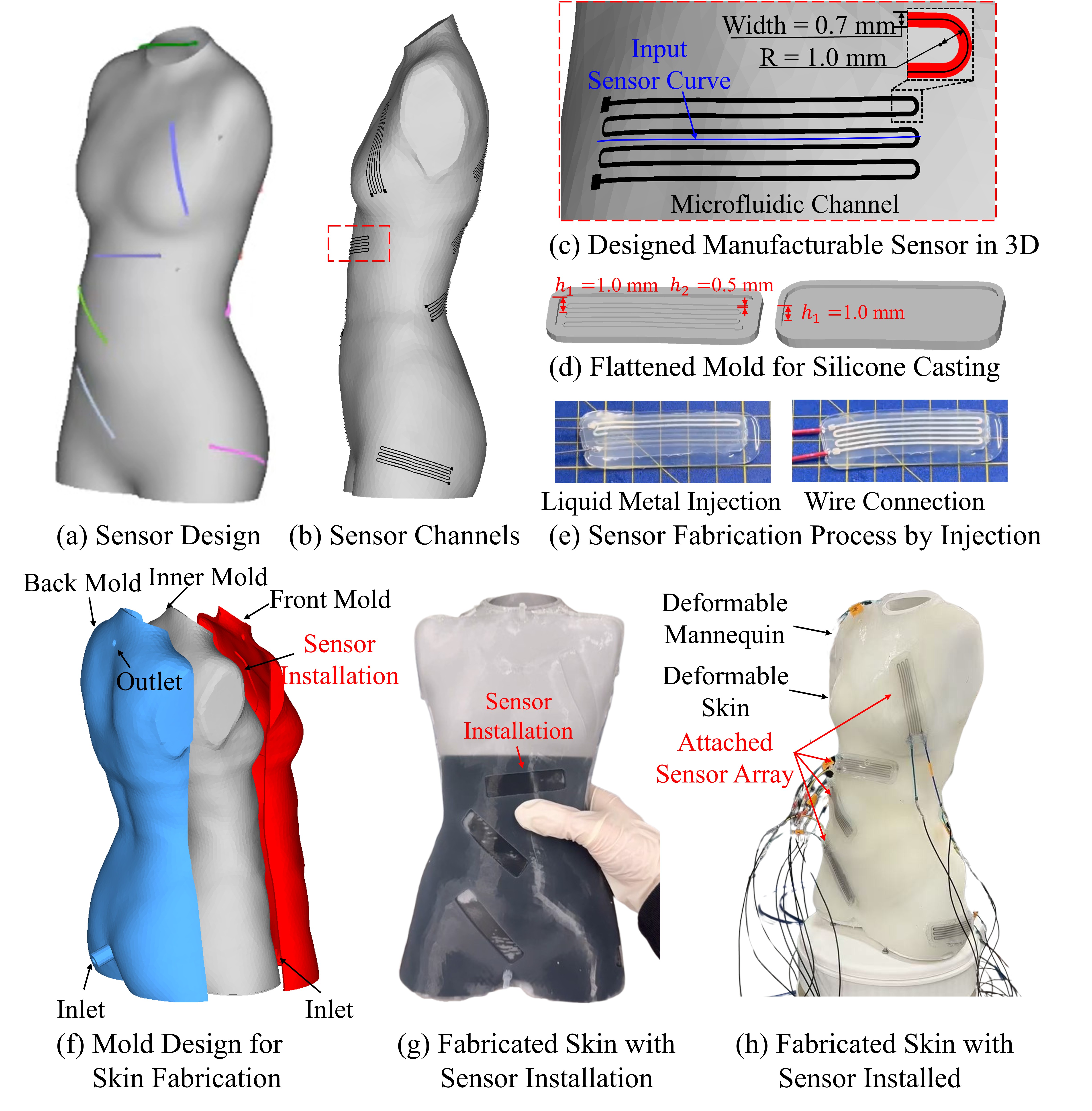}
\caption{Sensor and skin fabrication: (a) a sensor layout represented as a collection of curves, (b) offsetting curves to create sensor tracks, (c) connecting curves to form channels for liquid metal filling, (d) mold for sensor fabrication, (e) liquid metal injection and electrode wiring, (f) mold for skin fabrication, (g) mold with installation slots for placing sensors according to the optimized layout, and (h) final fabricated skin with integrated sensors. 
}\label{figSensorSkinFab}
\end{figure}

\subsection{Sensor \& skin fabrication}\label{subsecSensorFabCali}
The optimized layout of sensors are defined as a set of curves on a free-form surface, as depicted in Fig. \ref{figSensorSkinFab}(a). To fabricate these sensors, these curves are converted into conductive tracks by offsetting in a direction perpendicular to the curves -- see the illustration shown in Fig. \ref{figSensorSkinFab}(b) and the zoom-view in Fig. \ref{figSensorSkinFab}(c). Following the method described by \cite{dickey2017stretchable}, these tracks are converted into microfluidic channels filled with Galinstan -- a room-temperature liquid metal known for its high conductivity and low toxicity -- as the sensing material. The sensors with microfluidic channels are made in the shape of stretchable membranes by silicon (Ecoflex-0010), which are fabricated by the casting process (see Fig.\ref{figSensorSkinFab}(d)). 

The sensing mechanism relies on resistance variation caused primarily by changes in the sensor's length. This results in a highly linear and repeatable resistance response to deformation. Electrodes are connected to both ends of each sensor's microfluidic channel with liquid metal (Galinstan) injected for the resistance measurement, with the final configuration as shown in Fig. \ref{figSensorSkinFab}(e). Two 8-channel resistance reading devices are used to acquire sensor data, communicating with the host PC via the Modbus protocol. 

Stretchable sensors were integrated into the elastic skins of soft robots and deformable wearables. The skins were fabricated from silicone (Ecoflex-0030) via a molding process (Fig.~\ref{figSensorSkinFab}(f)), with the mold designed to incorporate predefined installation slots on the skin surface (Fig.~\ref{figSensorSkinFab}(g)). Following silicone injection and curing, the sensors were inserted into the molded slots to ensure accurate alignment with the optimized layout. An example of a fabricated skin with fully integrated sensors is shown in Fig.~\ref{figSensorSkinFab}(h).

\subsection{Network architecture} 
The shape prediction network, illustrated in Fig.~\ref{figPipeline}, is implemented as a multi-layer perceptron (MLP) with 3 hidden layers of 36 neurons each. \rev{This network architecture is chosen by empirical tests.} Batch normalization is applied to all layers, and ReLU is used as the activation function. 
To further aid learning, the network outputs offset vectors relative to the undeformed control points, enabling more efficient convergence (ref.~\cite{Bailey2020TOG}). The input signal $\mathbf{s}$ has a dimensionality of \rev{$M$} as the maximally allowed number of sensors, while the network outputs a vector of dimension $3mn$, corresponding to $mn$ B-spline control points in 3D space. We choose \rev{$M=20$} based on the consideration of fabrication time and system complexity. \rev{More details about the maximum number of sensors can be found in Sec.~\ref{subsec:Initialization}.}

\subsection{Dataset preparation}\label{subsecDataset}
Datasets are constructed for the three hardware setups described above. For each setup, ground-truth deformations are obtained using two steps: (1) capturing the positions of surface markers via a motion capture system, and (2) deforming the corresponding template models to fit the captured marker positions \rev{(ref.~\cite{kwok2014volumetric})}. B-spline surface control points are then generated by fitting surfaces to these deformed template models. For each setup, $2,000$ deformed shapes are collected \rev{by the preference to span a large space of shape variation} and processed into control point representations. \rev{Specifically, each shape is represented as a triangular mesh and then parameterized onto a square $(u,v)$ domain using the method of~\cite{floater2003mean}. The $m \times n$ control points of the corresponding B-spline surface are subsequently determined via a least-squares fitting, with an additional regularization term to encourage a nearly uniform distribution of control points. In detail,} the sensing shoulder example uses a resolution of $15 \times 15$ control points, while both soft robotic setups employ $30 \times 30$ control points. \rev{The number of control points is chosen by balancing the  shape approximation accuracy and the computational efficiency.} Each dataset is split into 80\% for training and 20\% for testing.

\begin{figure}[t]
\centering
\includegraphics[width=1.0\linewidth]{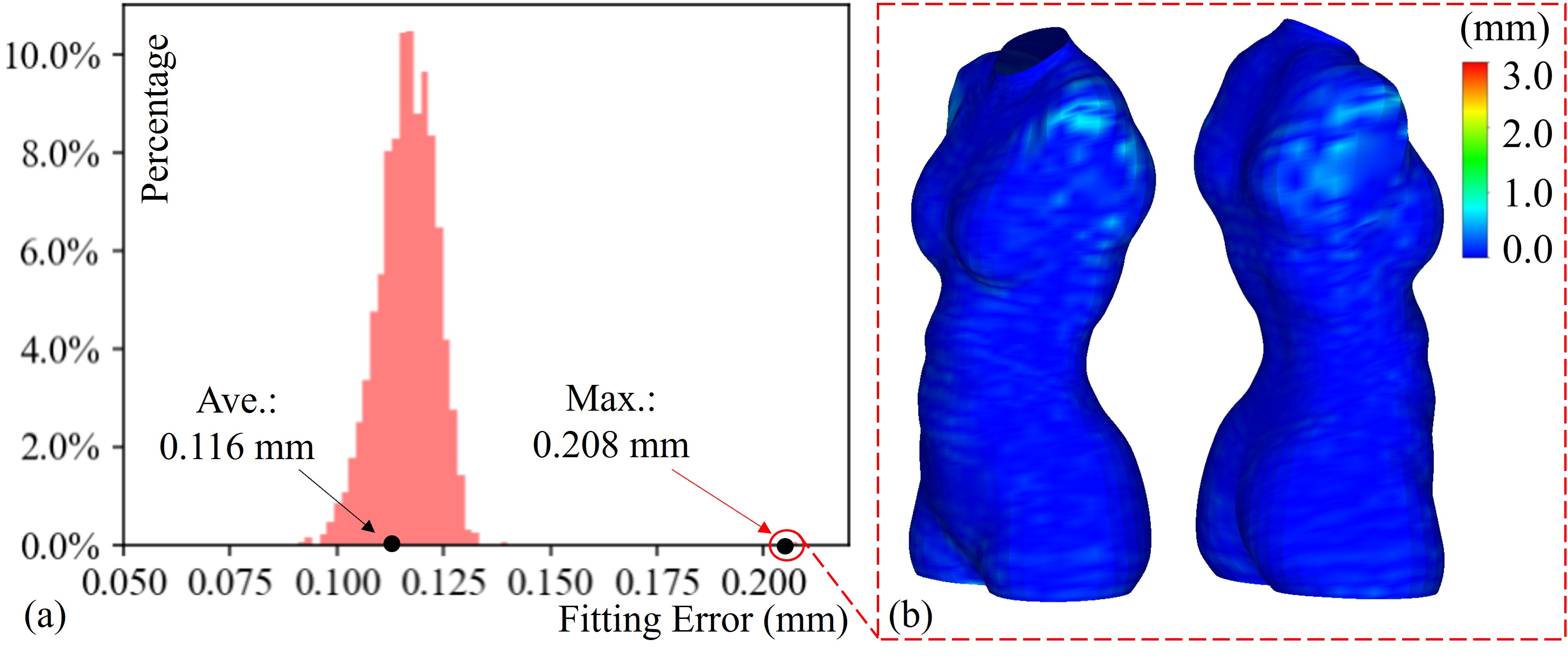}
\vspace{-15pt}
\caption{\rev{Distribution of the B-spline surface fitting errors for the dataset of deformable mannequin. While the mean error (0.116 mm) and the maximum error (0.208 mm) are indicated by the arrows, the distribution has a very small standard deviation (0.006 mm).}
}\label{figFittingErr}
\end{figure}

\rev{To verify the accuracy of the B-spline approximation, we compute the surface fitting errors for all 2,000 shapes in the deformable mannequin dataset, with the error distribution shown in Fig.~\ref{figFittingErr}. As observed, highly accurate surface fitting is achieved, with a mean error of 0.116mm and a small standard deviation of 0.006mm. Note that errors in sensor-based shape prediction may arise from non-uniform distortion during the mapping from the $(u,v)$-domain to Euclidean space. These effects are mitigated by the use of an advanced surface parameterization and a regularization term in the fitting process that promotes a nearly uniform distribution of B-spline control points.}

\subsection{Training and weights}\label{subsecWeightTuning}
The training of our MLP-based neural network and the associated parameterized sensor layout tensors is conducted using the Adam optimizer. A learning rate of 0.06 is chosen by experiments, with a maximum of 100 training epochs and a batch size of 16. Convergence curves from the training process are presented in the Results section.

The weights used in the formulation of the total loss (i.e., Eq.~\eqref{eqTotal}) are empirically determined through stepwise tuning \rev{as presented below}, following the strategy \rev{of giving highest priority to shape prediction function and then progressively adding other objectives and constraints}:
\begin{itemize}
\item The weight $w_t$ is tuned first, with $w_m=w_p=w_s=0.0$. It was observed that large values of $w_t$ lead to solutions with very few sensors. Starting from $w_t=1.0$, it was iteratively reduced by a factor of 0.1 until the resulting sensor count converged to approximately 20. The final value was chosen as the midpoint between two consecutive orders of magnitude satisfying this criterion, which yields $w_t=0.005$

\item With $w_t=0.005$ fixed, the weight $w_m$ is tuned to control the minimum sensor length. Beginning from $0.001$, we increment $w_m$ in logarithmic steps (i.e., $\times 10$ each step) until the sensor length can be effectively constrained. This results in $w_m=0.1$.
    
\item The weight $w_p$ is then adjusted to prevent sensor overlaps. With $w_t = 0.005$ and $w_m = 0.1$, we increase $w_p$ from zero using a step size of 0.1 until overlaps are effectively avoided, which gives $w_p=0.6$.

\item Finally, the weight $w_s$ is tuned to enforce the inter-sensor distance in sensor layout. We gradually increase $w_s$ from zero with a step size of 0.001, and obtain $w_s=0.005$ as a satisfactory value.
\end{itemize}
The above strategy is first applied to the soft robotic mannequin setup, and the determined weight values are found to perform effectively on the other two hardware configurations as well.

\begin{figure}[!t]
\centering
\includegraphics[width=1.0\linewidth]{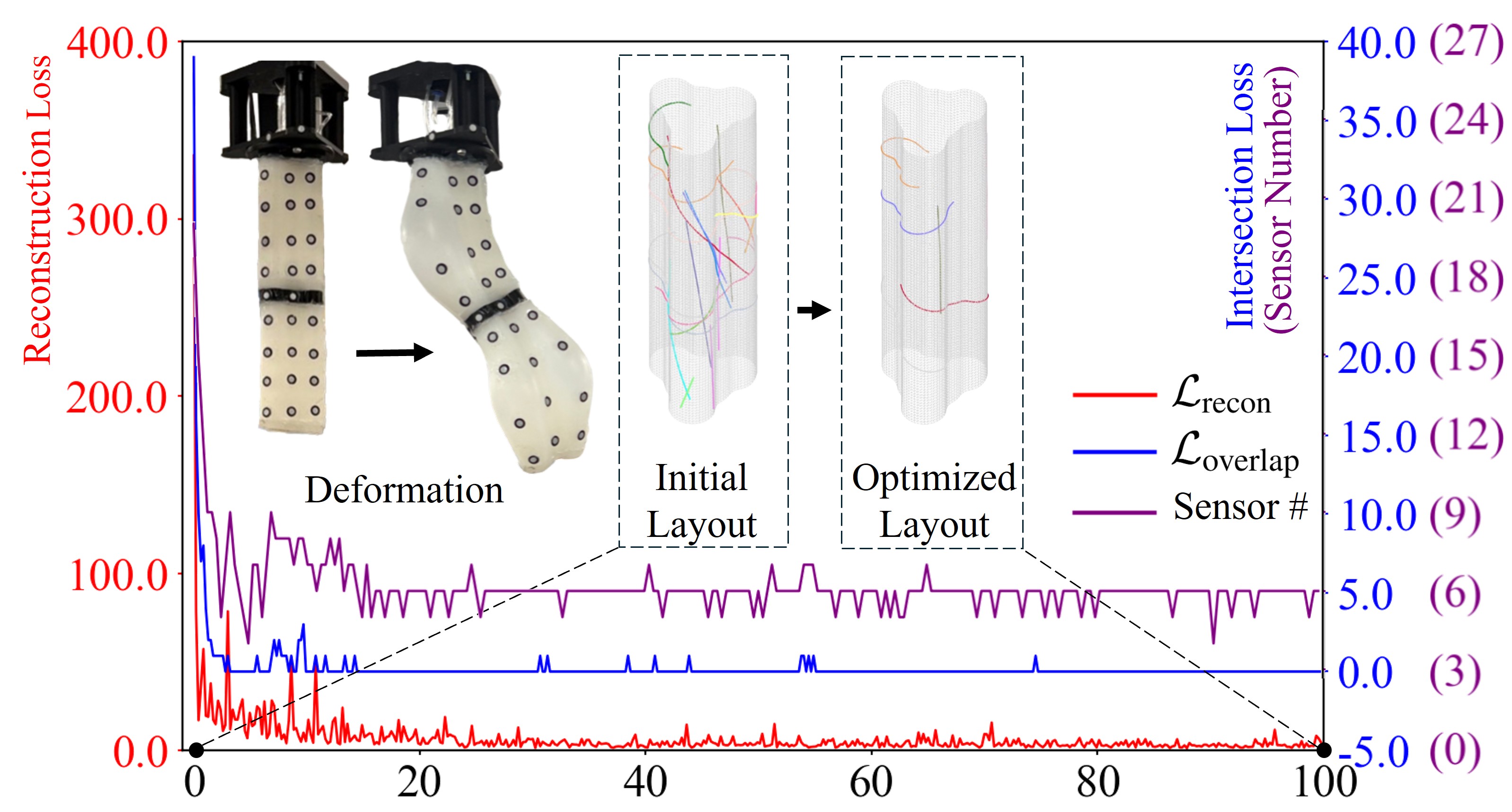}
\caption{Optimization process of sensor layout on the soft manipulator based on the dataset with deformation tracked by the markers, where the convergence curves of the reconstruction loss (in \textcolor{red}{red}), the overlap avoidance loss (in \textcolor{blue}{blue}), and the number of sensors (in \textcolor{mypurple}{purple}) are given. The initial and the optimized sensor layouts, corresponding to the starting and converged states of the optimization, are highlighted. 
}\label{figManipulatorExpInit}
\end{figure}

\begin{figure}[!t]
\centering
\includegraphics[width=0.95\linewidth]{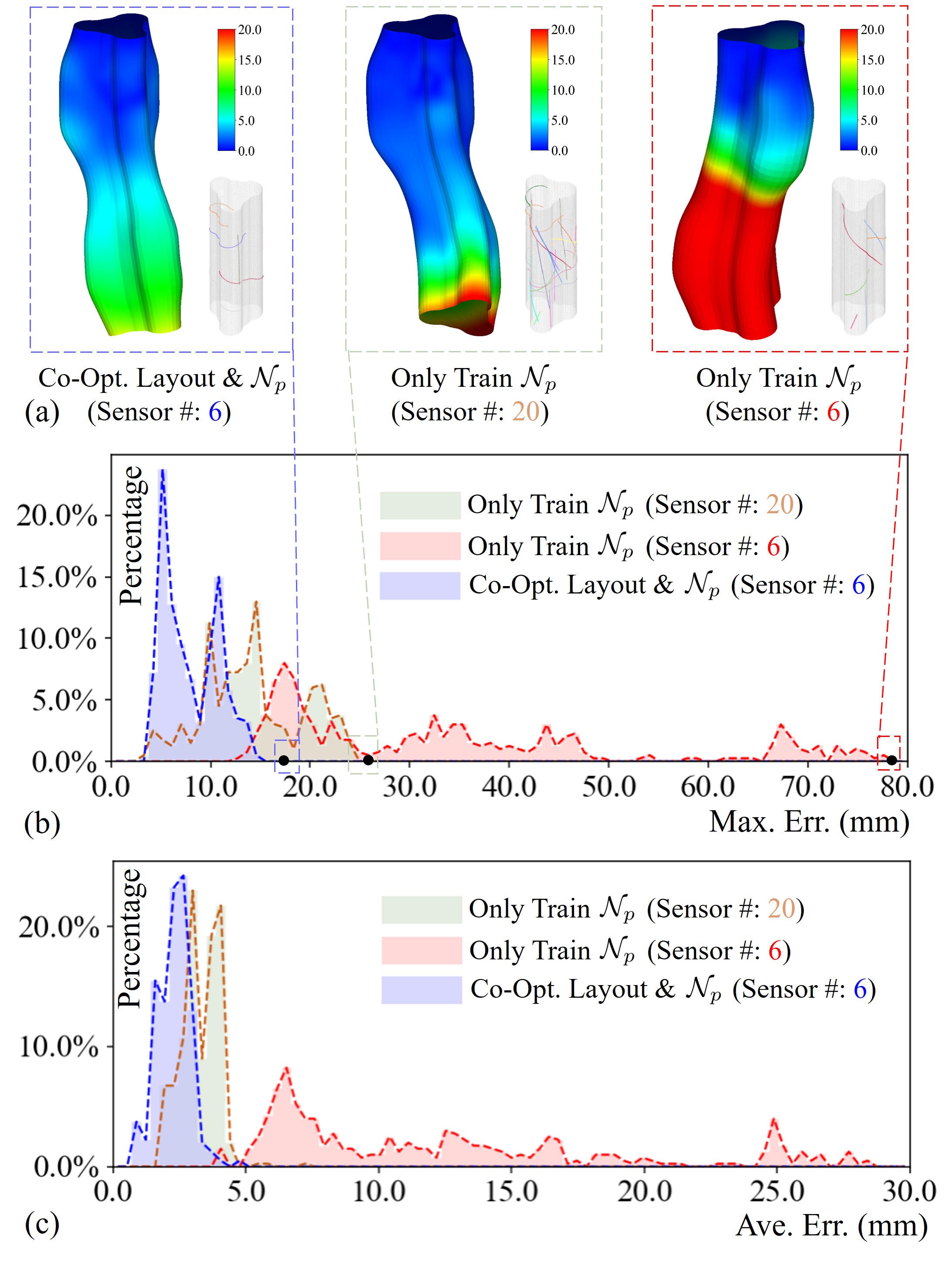}
\vspace{-10pt}
\caption{
Results of sensor layout optimization with 6 sensors, compared to unoptimized layouts with 20 and 6 sensors -- i.e., with only the shape prediction network $\mathcal{N}_p$ trained: (a) sensor layouts and color maps showing the approximation errors on the worst cases, (b) distributions of maximum errors over 400 test samples, and (c) distributions of average errors.
}\label{figRandomResultsManipulator}
\end{figure}

\begin{figure}[!t]
\centering
\includegraphics[width=.95\linewidth]{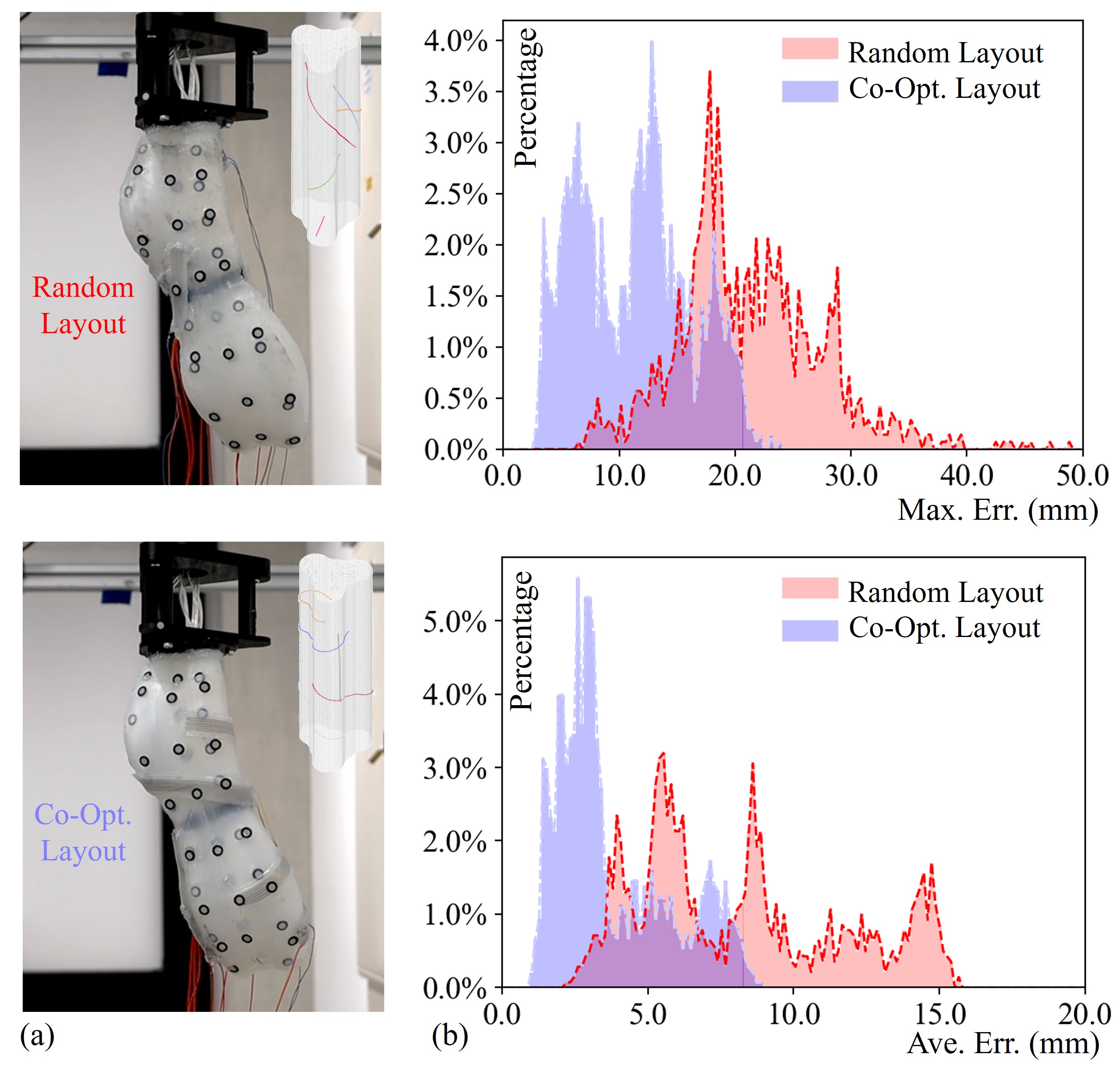}
\vspace{-10pt}
\caption{Physical experiments on soft manipulation comparing proprioception errors with optimized and unoptimized 6-sensor layouts: (a) sensor layouts and (b) distributions of maximum and average shape prediction errors.
}\label{figValidationManipulator}
\end{figure}

\begin{figure}[!t]
\centering
\includegraphics[width=\linewidth]{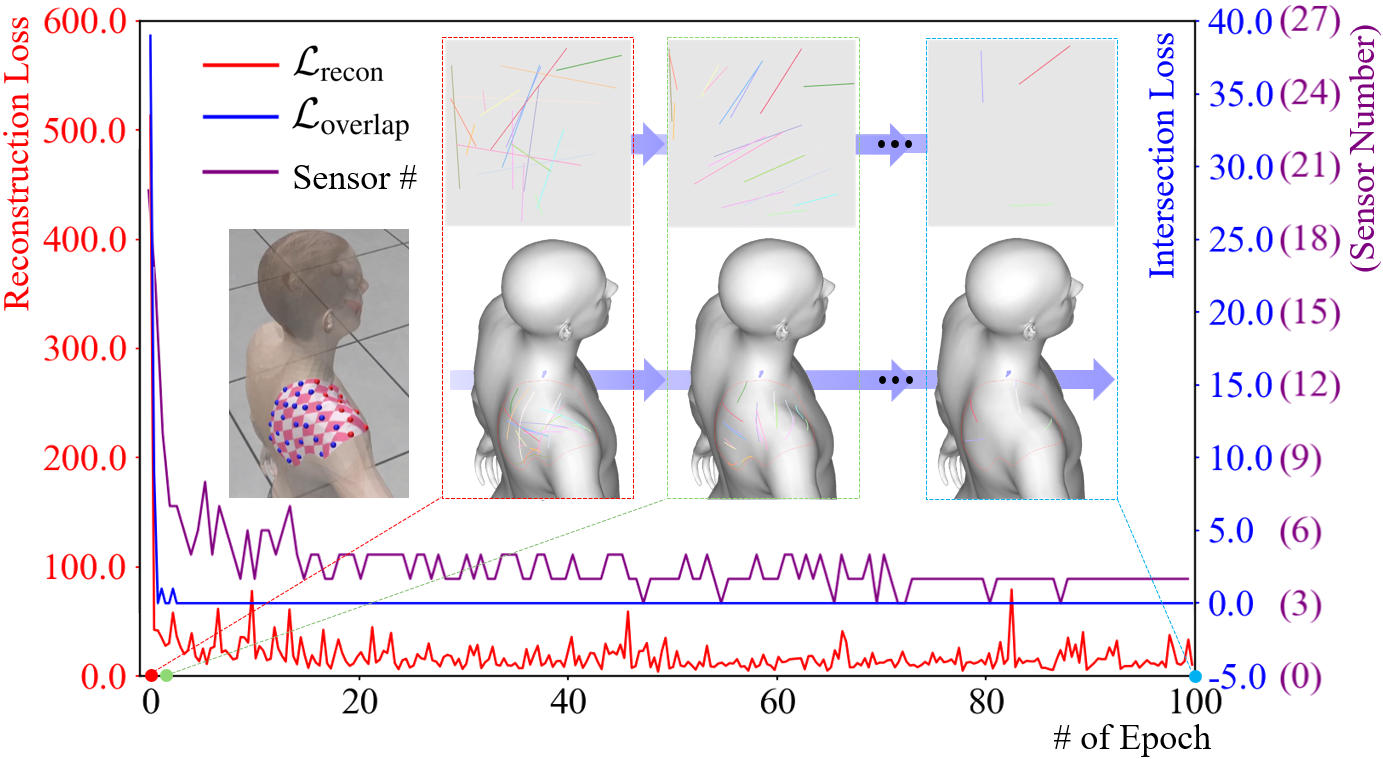}
\vspace{-20pt}
\caption{The co-optimization taken on a shoulder deformation sensing wearable, where the convergence curves of the reconstruction loss (in \textcolor{red}{red}), the overlap avoidance loss (in \textcolor{blue}{blue}), and the number of sensors (in \textcolor{mypurple}{purple}) are displayed.
}\label{figRandomIterationsShoulder}
\end{figure}

\section{Experimental Results}\label{secResult}
This section presents experimental results on the three hardware setups described above. Both computational and physical tests were performed to evaluate the advantages of optimized sensor layouts for deformation proprioception. In all examples, the ground-truth surface geometry was obtained by fitting B-spline surfaces to marker positions captured with a MoCap system, following a model-free, data-driven methodology.

\subsection{Soft manipulator}
The first example we tested is the soft manipulator. Shape differences under deformation were captured using markers, as shown in Fig.~\ref{figManipulatorExpInit}, and used to build the dataset described in Sec.~\ref{subsecDataset}. The co-optimization was performed on 80\% of the $2,000$ captured shapes, producing the optimized sensor layout as shown in the right of Fig.~\ref{figManipulatorExpInit}. The joint optimization of sensor layout and the shape prediction neural network converged within 153 minutes over 100 epochs. The process started from a random distribution of 20 long and intersecting sensors, but the results show that only 6 sensors are sufficient for accurate shape prediction on this soft manipulator with six pneumatic chambers.

The performance of the optimized layout was first evaluated on the test dataset (20\% of the $2,000$ captured shapes). Prediction results were compared with those obtained from unoptimized layouts of 20 and 6 sensors respectively (see Fig.~\ref{figRandomResultsManipulator}). The distributions of both average and maximum prediction errors clearly demonstrate that the proposed co-optimization method significantly improves the accuracy of proprioception. Representative cases with the largest prediction errors for different layouts are shown in the right of Fig.~\ref{figRandomResultsManipulator}(a). It is worth noting that the 20-sensor layout cannot be physically fabricated due to the manufacturing constraints discussed earlier in the paper.

Physical experiments were also conducted to compare two layouts of 6 sensors (see Fig.~\ref{figValidationManipulator})\rev{, where the unoptimized layout was randomly generated by eliminating the random layout with sensor-intersection}. The results confirm that the optimized layout substantially reduces shape prediction errors compared to the unoptimized counterpart.

\begin{figure}[!t]
\centering
\includegraphics[width=\linewidth]{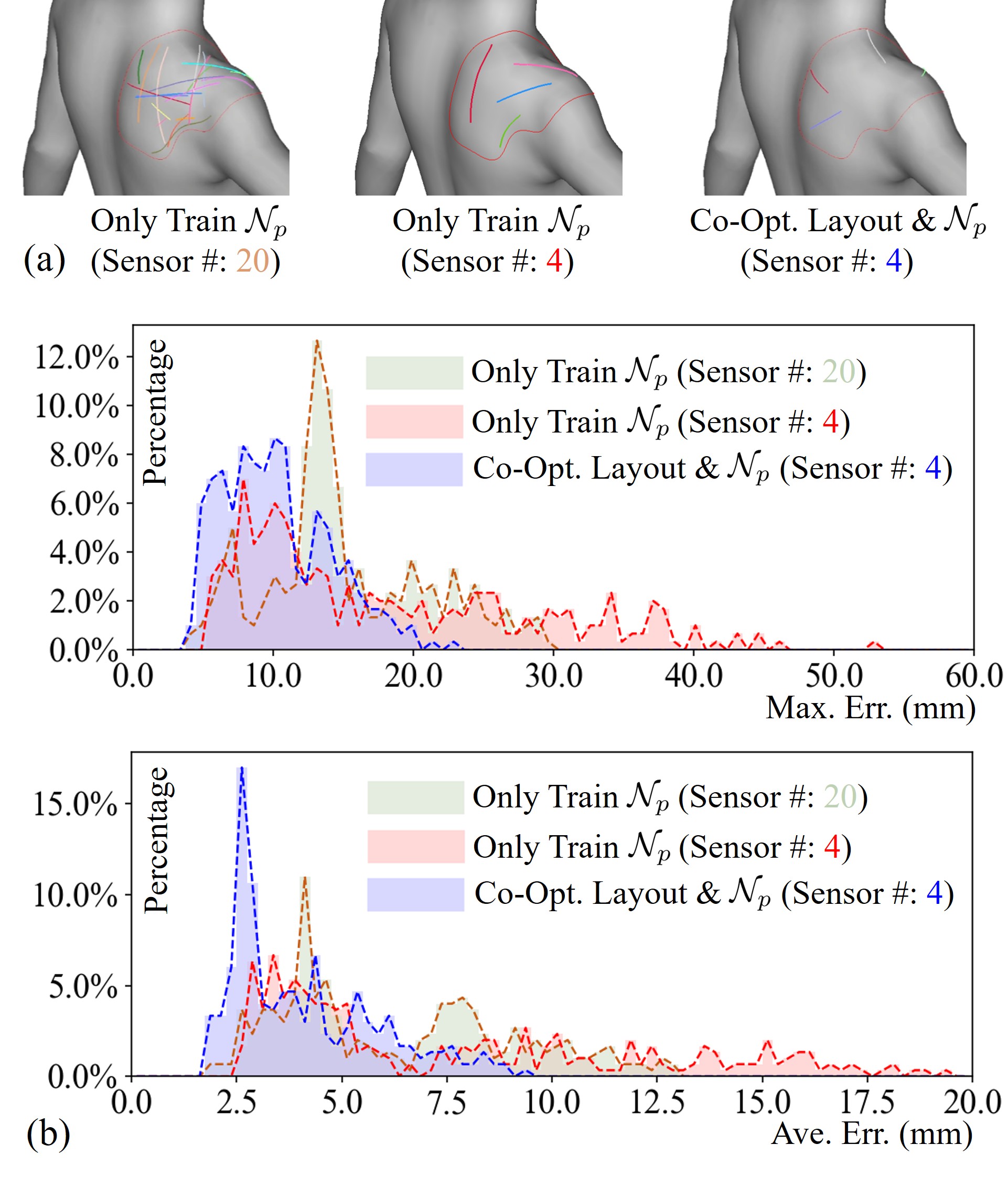}
\vspace{-15pt}
\caption{
Results of sensor layout optimization with 4 sensors, compared to unoptimized layouts with 20 and 4 sensors -- i.e., with only the shape prediction network $\mathcal{N}_p$ trained: (a) sensor layouts and (b) distributions of average and maximum errors over 400 test samples.
}\label{figRandomResultsShoulder}
\end{figure}

\begin{figure}[!t]
\centering
\includegraphics[width=\linewidth]{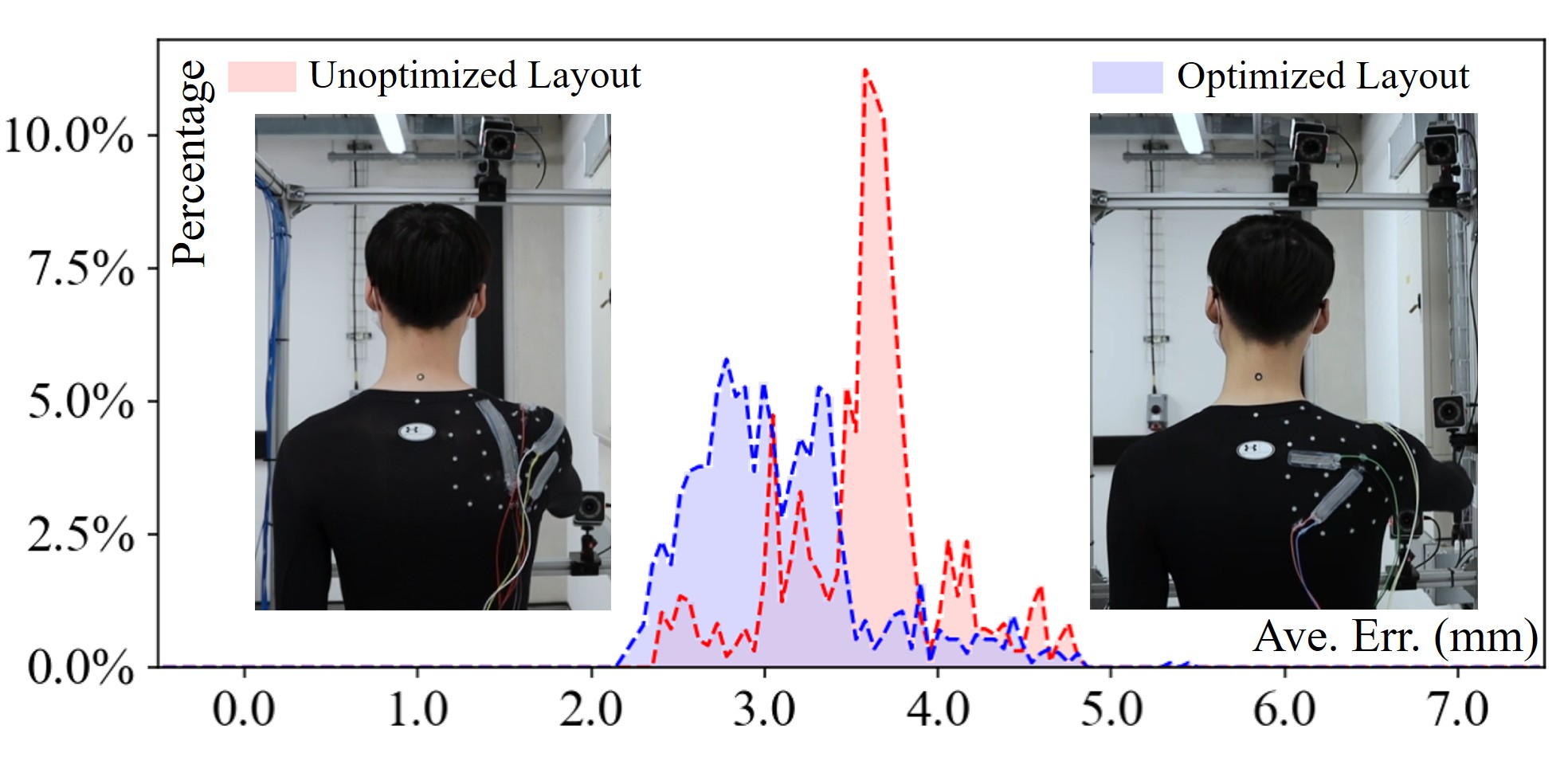}
\vspace{-15pt}
\caption{Comparison of deformation sensing during shoulder motion using an unoptimized sensor layout vs. an optimized sensor layout -- both have 4 sensors. Shape prediction errors were evaluated over $1,000$ poses, with ground-truth deformations captured by MoCap. 
}\label{figValidationShoulder}
\end{figure}

\subsection{Deformation sensing wearable}
For the experimental tests on the deformation-sensing shoulder wearable, we first selected a template mesh from the SMPL-X dataset \cite{SMPL-X:2019} that matched with the subject’s body height and weight. The template was then deformed to fit the body shape captured by motion-capture markers, using volumetric deformation techniques (e.g.,~\cite{kwok2014volumetric}). B-spline surfaces were subsequently constructed for the region of interest and fit to the marker positions, producing a dataset of $2,000$ shapes for training and testing.

The co-optimization of the sensor layout $(\mathbf{L},\mathbf{b})$ and the shape prediction network $\mathcal{N}_p$ was performed on 80\% of the dataset, with the convergence process shown in Fig.~\ref{figRandomIterationsShoulder}. Starting from a random distribution of 20 long, intersecting sensors, the optimization converged to a compact layout of 4 sensors, which proved sufficient for accurate shape prediction on the wearable. The process completed within 36 minutes over 100 epochs.

The optimized layout was then compared with results obtained from training only the prediction network $\mathcal{N}_p$ on unoptimized layouts. As shown in Fig.~\ref{figRandomResultsShoulder}, the co-optimized layout with 4 sensors significantly outperforms unoptimized layouts with both 20 and 4 sensors. Physical experiments were further conducted on fabricated wearables (Fig.\ref{figValidationShoulder}), and the results confirm a substantial improvement in prediction accuracy when using the sensor layout optimized by our method.


\begin{figure*}
\centering
\includegraphics[width=\linewidth]{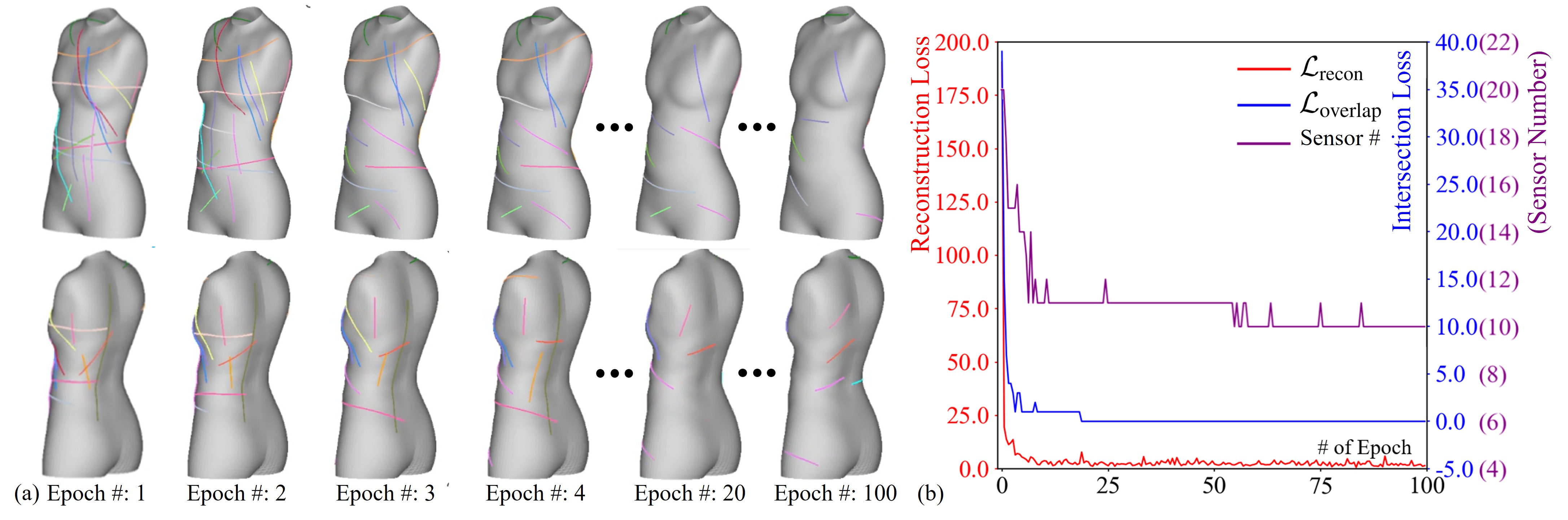}
\vspace{-20pt}
\caption{
Progressive results of sensor layout optimization starting from a random layout of 20 sensors, converging to a final layout of 10 short, spaced, and non-intersecting sensors. The right panel shows the convergence curves of the reconstruction loss (\textcolor{red}{red}), overlap avoidance loss (\textcolor{blue}{blue}), and the number of sensors (\textcolor{mypurple}{purple}).
}\label{figRandomIterations}
\end{figure*}

\begin{figure*}
\centering
\includegraphics[width=0.95\linewidth]{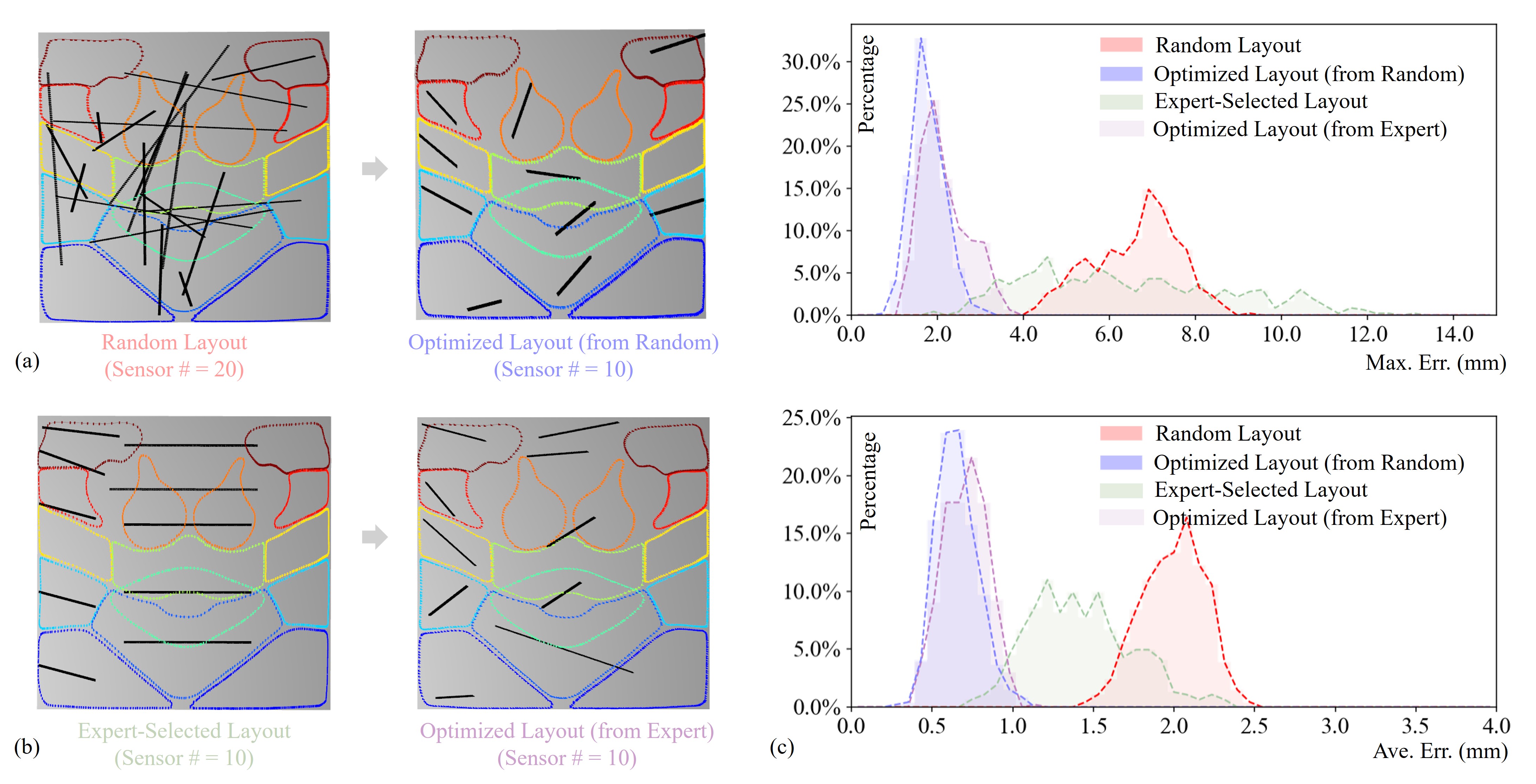}
\vspace{-10pt}
\caption{Results of sensor optimization on the deformable mannequin in its UV-domain (a) from a random initial layout with 20 sensors and (b) from an expert-designed initial layout with 10 sensors. The distributions of corresponding shape prediction errors are given in (c).
}\label{figExpertResults}
\end{figure*}

\subsection{Deformable mannequin}
We present the progressive optimization results of sensor layout for the deformable mannequin, starting from a random configuration of 20 sensors. The co-optimization process performed by our computational pipeline gradually transforms this layout into 10 short, well-spaced, and intersection-free sensors, while significantly reducing the shape prediction loss $\mathcal{L}_{recon}$ (Fig.~\ref{figRandomIterations}). The initial layout included 39 intersections among 20 sensors. Through optimization, the reconstruction loss was reduced from 175.1 to 2.5, the number of intersections decreased from 39 to 0, and the sensor count dropped from 20 to 10. Notably, all intersections were eliminated after only 20 epochs. The whole optimization process takes 197 minutes for 100 epochs .

The layouts in the UV-domain, both before and after optimization, along with their relative positions with respect to the pneumatic chambers, are shown in Fig.~\ref{figExpertResults}(a). The corresponding distributions of shape prediction errors are provided in Fig.~\ref{figExpertResults}(c).

To further evaluate our method, we compared it against a layout of 10 sensors designed by experts from the garment industry based on fabrication rules (see the left of Fig.~\ref{figExpertResults}(b)). The performance of this expert-designed layout in shape prediction is also reported in Fig.~\ref{figExpertResults}(c). We then applied our optimization framework to further optimize the layout, which produced a different 10-sensor layout but achieved comparable prediction accuracy, as shown in Fig.~\ref{figExpertResults}(c).

Finally, experimental verification was carried out on physically fabricated deformable mannequins equipped with 10 sensors, tested across $3,000$ poses. The error comparisons of shape prediction are summarized in Fig.~\ref{figTeaser}. These results show that our optimized layout with 10 sensors provides substantially more accurate proprioception than the expert-designed layout \rev{with the same number of sensors}.

\begin{table}[!t]\footnotesize
\color{revcolor} 
\vspace{0pt}
\caption{{Study of Sensor Layout Initialization}}
\label{Tab:sensor_opt_30Sensors}
\begin{tabular}{c|cc|cc|cc}
\hline 
\multicolumn{1}{c|}{{\textbf{Layout}}} &
\multicolumn{2}{c|}{{\textbf{Sensor Num.}}} &
\multicolumn{2}{c|}{{\textbf{Total Length}}$^\dagger$} &
\multicolumn{2}{c}{{\textbf{Max Shape Err.}}$^\ddagger$}
\\
\cline{2-7}

\multicolumn{1}{c|}{{\textbf{Trial ID}}} &
\multicolumn{1}{c|}{{Before}} &
\multicolumn{1}{c|}{{After}} &
\multicolumn{1}{c|}{{Before}} &
\multicolumn{1}{c|}{{After}} &
\multicolumn{1}{c|}{{Before}} &
\multicolumn{1}{c}{{After}}
\\

\hline 
1  & 20 & 10 & 3018.58    & 610.09   & 2.44   & 1.07   \\
2  & 20 & 10 & 3460.06    & 666.04   & 3.10   & 1.11   \\
3  & 20 & 10 & 3911.56    & 674.36   & 5.33   & 1.28   \\
4  & 20 & 11 & 3910.27    & 718.60   & 3.21   & 1.43   \\
5 & 20 & 10 & 3149.73    & 624.34   & 2.82   & 0.96   \\

\hline 
Mean      & -- & 10.2 & 3490.04 & 658.69 & 3.38 & 1.17 \\
Std. Dev. & -- & 0.45 & 416.31 & 43.10 & 1.13 & 0.19 \\

\hline 
\hline 

6 & 30 & 10 & 5969.71 & 717.52 & 2.65 & 1.21 \\
7 & 30 & 10 & 5188.21 & 604.75 & 2.68 & 0.87 \\
8 & 30 & 10 & 5393.14 & 605.43 & 6.20 & 1.19 \\
9 & 30 & 10 & 5870.43 & 641.98 & 2.66 & 1.20 \\
10 & 30 & 10 & 4587.68 & 627.52 & 3.62 & 0.70 \\

\hline

Mean & -- & 10.0 & 5401.83 & 639.44 & 3.56 & 1.03 \\
Std. Dev. & -- & 0.0  & 559.16  & 46.38 & 1.53 & 0.24 \\

\hline
\end{tabular}
\begin{flushleft}\footnotesize
$^\dagger$~Length Unit: mm. \\
$^\ddagger$~Maximum average surface distance error (Unit: mm).
\end{flushleft}
\vspace{-15pt}
\end{table}

\subsection{\rev{Influence of initial layout}}\label{subsec:Initialization}
\rev{In above subsection, we evaluated the performance of our sensor optimizer on the deformable mannequin dataset using two different initial configurations: (i) a layout with 20 sensors placed at randomly selected locations, and (ii) an expert-specified layout with 10 sensors used as the initial guess. As shown in Fig.~\ref{figExpertResults}, the optimization converges to different sensor layouts while achieving comparable shape prediction accuracy, in terms of both maximum and average errors.}

\rev{To further examine the robustness of our approach, we conducted additional experiments using five different random initial layouts with 20 sensors and five more random initial layouts with 30 sensors. In all cases, the optimization converges to a similar number of sensors with comparable total sensor lengths; statistical results are summarized in Table~\ref{Tab:sensor_opt_30Sensors}. We further evaluate the resulting sensor configurations on all 400 shapes in the test dataset, reporting the maximum error across the dataset. The shape prediction error is measured as the average surface distance between the predicted shape and the ground truth. As shown in the results, similar prediction errors are consistently obtained, with small standard deviations.}

\rev{Note that, for all physical experiments conducted on three hardware setups presented above, we always choose the layout with the smallest shape prediction error among multiple trials to conduct the physical fabrication of sensor.}

\begin{figure*}[t]
\centering
\includegraphics[width=\linewidth]{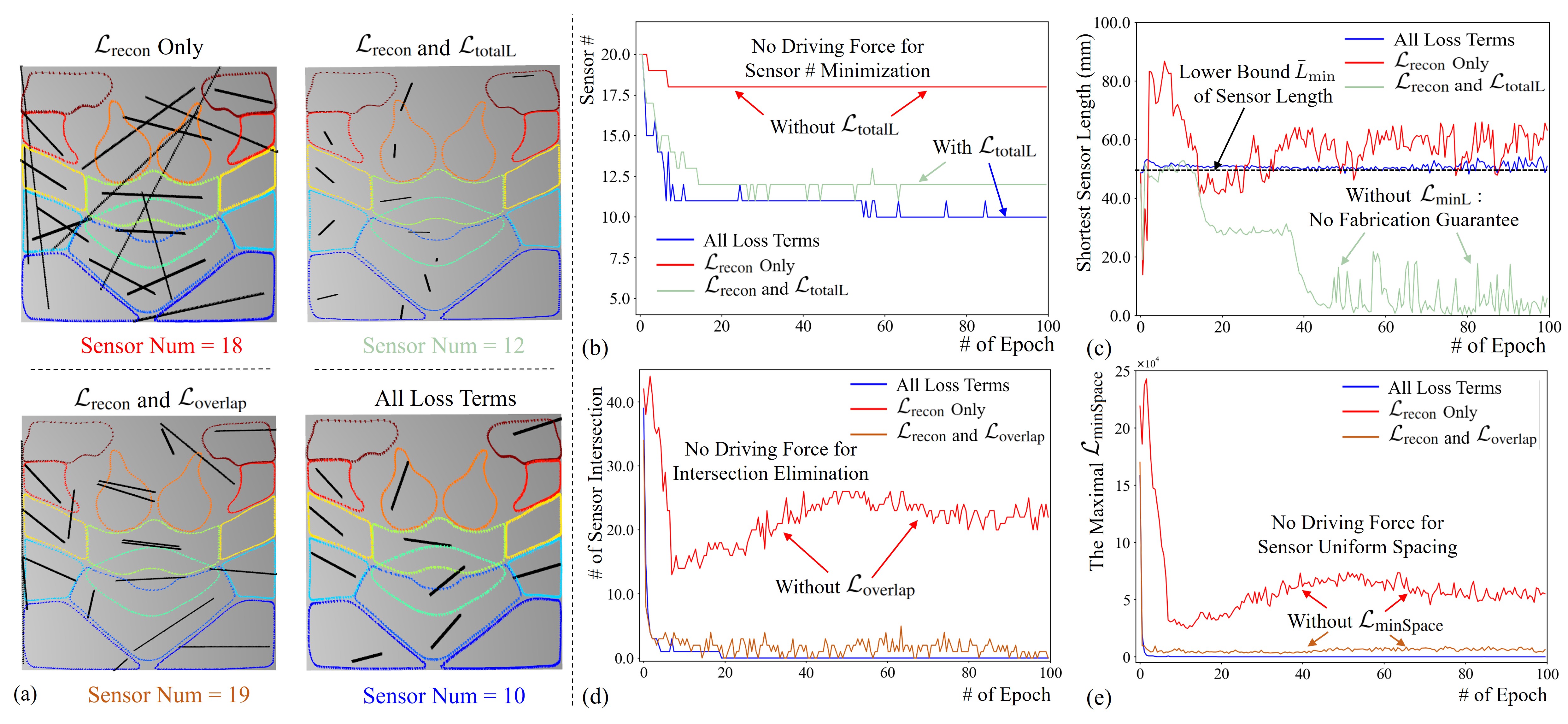}
\vspace{-20pt}
\caption{Ablation studies validating the necessity of each loss term. (a) Sensor layouts obtained with different combinations of loss terms. Convergence curves for these combinations are shown for (b) number of sensors, (c) length of the shortest sensor, (d) number of intersections, and (e) the maximum value of $\mathcal{L}_{\text{minSpace}}$ across all sensor pairs. 
}\label{figLossResults}
\end{figure*}

\subsection{Ablation study of loss functions}
We conducted ablation studies to validate the necessity of each loss term. Figure~\ref{figLossResults} shows the resulting sensor layouts and the corresponding convergence curves for different evaluation metrics. As can be found that using only the reconstruction loss $\mathcal{L}_{\text{recon}}$ (with all other terms removed) fails to enforce manufacturability -- i.e., the optimization produces numerous long, intersecting sensors at the end of computation. This case serves as the baseline for comparison.

Adding the total length loss $\mathcal{L}_{\text{totalL}}$ reduces overall sensor length but results in many extremely short sensors that are not manufacturable. The convergence curves show that \rev{$\mathcal{L}_{\text{totalL}}$ and $\mathcal{L}_{\text{minL}}$ work together to help indirectly} control the number of sensors, while the shortest sensor length can be effectively regulated by introducing the minimum length loss $\mathcal{L}_{\text{minL}}$ -- see the resultant curve without $\mathcal{L}_{\text{minL}}$ in Fig.~\ref{figLossResults}(c).

\begin{figure}[t]
\centering
\includegraphics[width=0.92\linewidth]{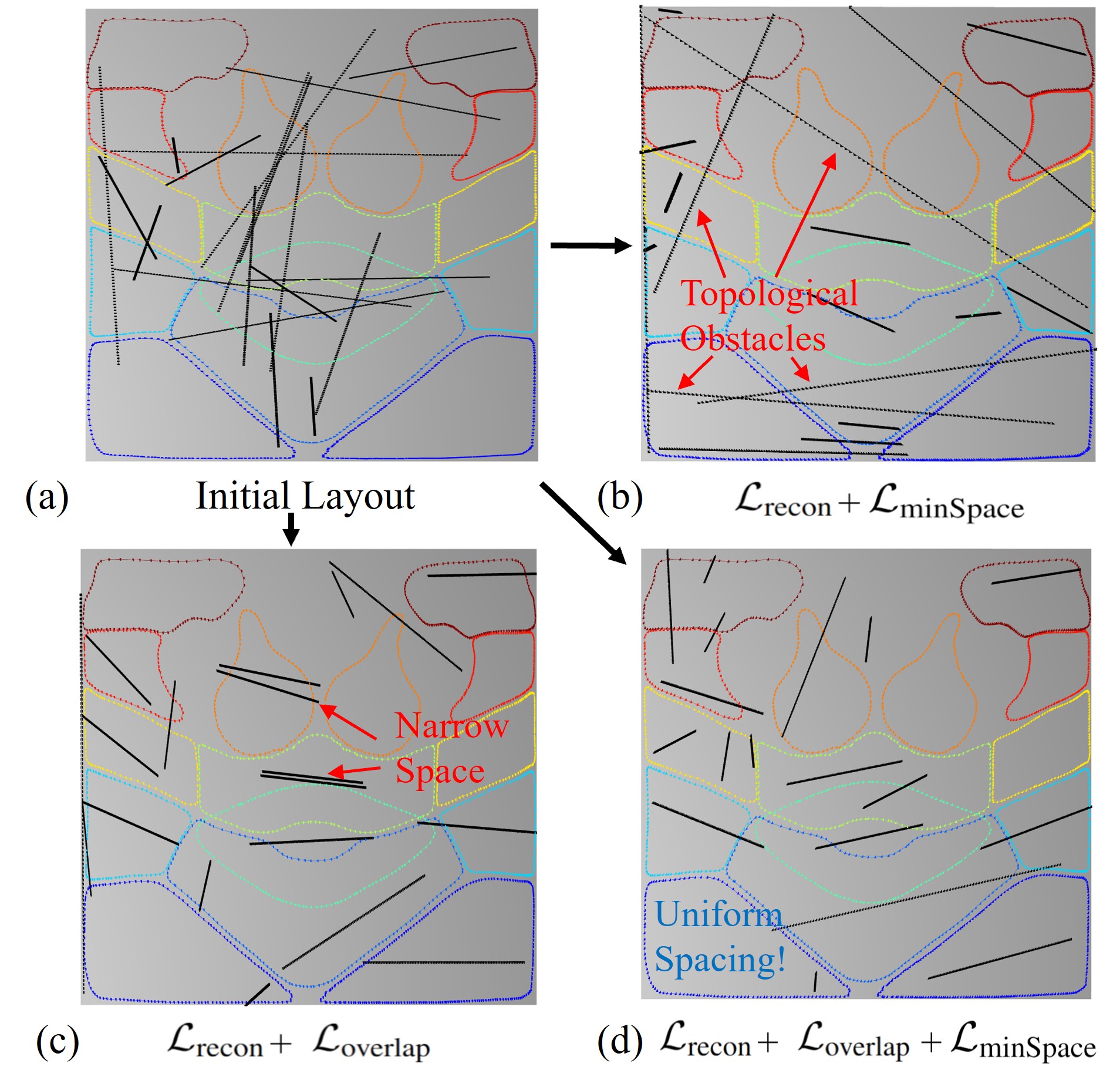}
\vspace{-10pt}
\caption{\rev{Ablation study of $\mathcal{L}_{\text{minSpace}}$ and $\mathcal{L}_{\text{overlap}}$: (a) an initial random layout with 20 sensors, (b) the result with only $\mathcal{L}_{\text{recon}}$ and $\mathcal{L}_{\text{minSpace}}$ where intersection cannot be completely removed due to topological obstacles, (c) the result with only $\mathcal{L}_{\text{recon}}$ and $\mathcal{L}_{\text{overlap}}$ shows no intersection, and (d) the result by using both $\mathcal{L}_{\text{overlap}}$ and $\mathcal{L}_{\text{minSpace}}$ together with $\mathcal{L}_{\text{recon}}$.}
}\label{figMinimalSpacingAbaltion}
\end{figure}

On the other hand, combining $\mathcal{L}_{\text{recon}}$ with the overlap loss $\mathcal{L}_{\text{overlap}}$ effectively prevents sensor intersections but neighboring sensors remain too close (see Fig.\ref{figLossResults}(d) \rev{and Fig.\ref{figMinimalSpacingAbaltion}(c)}). This highlights the importance of the inter-sensor distance loss $\mathcal{L}_{\text{minSpace}}$, whose influence is evident in the convergence curves of the minimum inter-sensor spacing (see Fig.~\ref{figLossResults}(e)). \rev{However, minimizing the $\mathcal{L}_{\text{minSpace}}$ loss alone does not guarantee an intersection-free sensor layout. As shown in Fig.~\ref{figMinimalSpacingAbaltion}(b), topological obstacles may arise -- e.g., two intersected excessively long sensors, which prevent the formation of intersection-free layouts when only $\mathcal{L}_{\text{recon}}$ and $\mathcal{L}_{\text{minSpace}}$ are used.}

\begin{figure*}[t]
\centering
\includegraphics[width=0.99\linewidth]{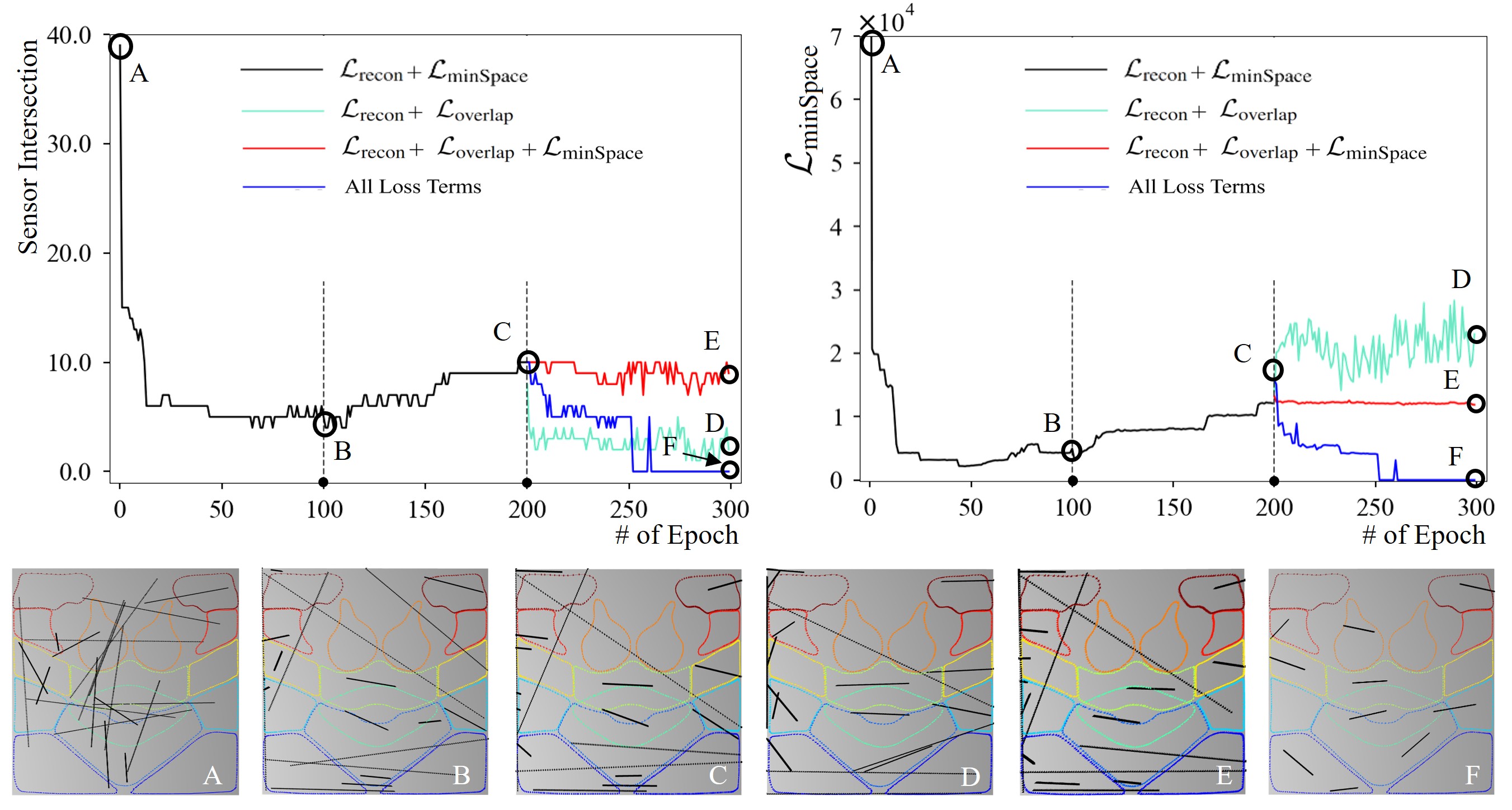}
\caption{\rev{When only $\mathcal{L}_{\text{minSpace}}$ is used together with $\mathcal{L}_{\text{recon}}$, the optimization becomes trapped by topological obstacles, manifested as two excessively long sensors, as shown in (B) and (C) after 100 and 200 epochs, respectively. Starting from (C), the optimization cannot escape this local minimum, either by (i) switching from $\mathcal{L}_{\text{minSpace}}$ to $\mathcal{L}_{\text{overlap}}$ (light-blue optimization path, resulting in (D)) or by (ii) adding $\mathcal{L}_{\text{overlap}}$ to work together with $\mathcal{L}_{\text{minSpace}}$ (red optimization path, resulting in (E)). In contrast, when all loss terms are applied from the beginning, the optimizer successfully converges to an intersection-free layout, as shown in (F).}
}\label{figLossAbalationStudy2}
\end{figure*}

\rev{Another interesting study is to further optimize the layout with topological obstacle (i.e., as shown in Fig.\ref{figMinimalSpacingAbaltion}(b)) by using different combinations of loss terms, such as:}
\begin{enumerate}
    \item \rev{switching from $\mathcal{L}_{\text{minSpace}}$ to $\mathcal{L}_{\text{overlap}}$;}

    \item \rev{adding $\mathcal{L}_{\text{overlap}}$ to work together with $\mathcal{L}_{\text{minSpace}}$;}

    \item \rev{using all loss terms in $\mathcal{L}_\text{total}$.}
\end{enumerate}
\rev{The results are shown in Fig.~\ref{figLossAbalationStudy2}. We observe that the optimization process can escape the topological obstacle only after introducing the length-control losses $\mathcal{L}_{\text{minL}}$ and $\mathcal{L}_{\text{totalL}}$, which explicitly encourage shorter sensor lengths. In contrast, as illustrated in Fig.~\ref{figMinimalSpacingAbaltion}(d), the optimization is less prone to falling into scenarios with topological obstacles when both $\mathcal{L}_{\text{overlap}}$ and $\mathcal{L}_{\text{minSpace}}$ are applied from the very beginning.}

\begin{figure*}[t]
\centering
\includegraphics[width=.92\linewidth]{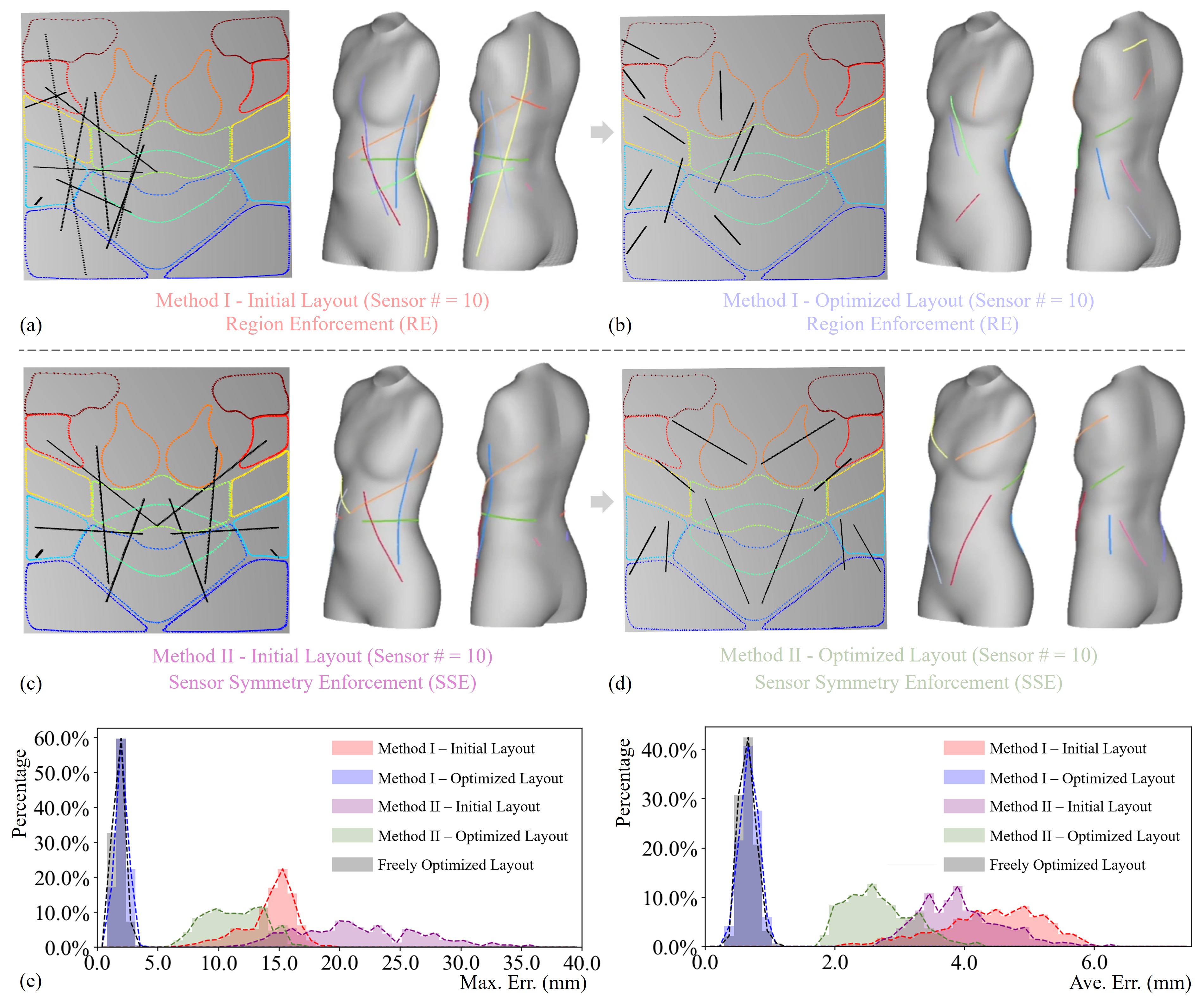}
\vspace{-15pt}
\caption{Study on whether symmetry in sensor layout needs to be explicitly enforced. (a, b) Method I: sensors restricted to one side. (c, d) Method II: symmetric distribution enforced. (e) Maximum and average shape prediction errors over 400 test shapes for layouts generated by these two methods and the free optimization method.
}\label{figSymmetryResults}
\end{figure*}

\subsection{Symmetric vs. asymmetric}
When sensor optimization is applied to a deformable system with symmetric deformation patterns, it is important to study whether our co-optimization pipeline can effectively capture such symmetry in shape prediction. We explored two different approaches.

In the first method, sensors were restricted to one side of the design space -- specifically to keep $(u,v)$ values within $[0, 0.5] \times [0, 1.0]$, as shown in Fig.~\ref{figSymmetryResults}(a). In the second method, we explicitly enforced symmetry by constraining 5 pairs of sensors to be mirrored about $u=0.5$, ensuring that their endpoints were symmetric in $u$ (Fig.~\ref{figSymmetryResults}(b)).

The sensor layouts generated by these two methods were tested on 400 sample shapes and compared with the layout obtained from free optimization (Fig.\ref{figRandomIterations}). The error distributions in Fig.\ref{figSymmetryResults}(c) show that Method I achieves performance comparable to free optimization. In contrast, Method II underperforms, as only half of the sensor signals are effectively used for shape prediction. 
\rev{All models in the dataset are nearly but not perfectly symmetric. These experiments demonstrate that in such nearly symmetric cases, the NN-based shape prediction can accurately reconstruct surface geometry when the sensors are restricted to only one side of the models. This provides more design options for sensor layout.}

\section{Conclusion and Discussion}\label{secConclusion}

In this paper, we addressed the challenge of optimal sensor placement for deformable freeform surfaces by introducing a model-free, data-driven computational pipeline that jointly optimizes the layout of flexible length-measurement sensors and the parameters of a shape prediction network. Unlike heuristic or trial-and-error approaches, our method explicitly integrates prediction accuracy with manufacturability constraints through differentiable loss functions for overlap avoidance, inter-sensor spacing, and length control (both total and minimum). To the best of our knowledge, this is the first pipeline that co-optimizes continuously represented sensor layouts together with a neural network for deformation proprioception. Furthermore, the method relies solely on datasets of deformed shapes, making it broadly applicable to diverse robotic sensing tasks without the need for \rev{physical simulation models}.

The effectiveness and generality of our approach have been validated across three hardware platforms: a soft deformable mannequin, a soft manipulator, and a shoulder deformation-sensing wearable. Results consistently show that, regardless of whether the initial layout is expert-designed or randomly generated, our method achieves superior shape reconstruction accuracy while eliminating issues such as redundant sensors, intersections, and narrow spacing. Beyond improving accuracy, the optimized layouts capture underlying deformation patterns (such as symmetry) beyond conventional human intuition, enabling accurate proprioception of large 3D deformations under practical fabrication constraints. In summary, this work introduces a promising direction for the computational design of sensorized systems, advancing both the theory and practice of proprioception in soft robotics and wearable devices.

Although the experimental results are promising, our method has several limitations as discussed below.
\begin{itemize}
    \item First, the current pipeline relies on datasets represented as B-spline surfaces, which requires dedicated pre-processing to convert captured point clouds or meshes into B-spline surfaces through parameterization and fitting. In future work, we plan to relax this requirement by directly handling point clouds and meshes, thereby lowering the barrier for applying our pipeline.

    \item Second, the design domain is currently restricted to the UV space $[0,1.0] \times [0,1.0]$, necessitating a cutting operation to flatten the input mesh into this domain. This introduces additional preprocessing effort and limits optimization, as sensors cannot cross design boundaries. To address this, we plan to integrate periodic parameterization, enabling sensor parameters to transition seamlessly between boundary values (near 0.0 and 1.0), which would increase the flexibility of sensor layout design.

    \item Finally, our current work mainly focuses on deformation-based shape reconstruction and does not account for collision or self-collision. Extending the framework to incorporate these aspects, by modeling richer signals and predicting contact scenarios through sensor design and optimization, represents an important future research.
\end{itemize}
All these limitations will be investigated in the future together with the new sensor fabrication techniques such as coaxial extrusion (\cite{Ames2025CoreShell,Kjar2024Coaxial}) and multi-axis 3D printing (\cite{FANG2024AMCCF,HONG2023MultiAxis}).

\section*{Acknowledgement}
The project is supported by the chair professorship fund of the University of Manchester and the research fund of UK Engineering and Physical Sciences Research Council (EPSRC) (Ref.\#: EP/W024985/1). Guoxin Fang was partially supported by the Hong Kong SAR Research Grants Council Early Career Scheme (RGC-ECS) (CUHK/24204924).

\bibliographystyle{IEEEtran}
\bibliography{TRO.bib}

\section*{Appendix A}
The gradient of $L_{s}$ w.r.t. the sensor parameters (i.e., $u_s, v_s, u_e, v_e$) are defined as follows:
\begin{flalign}
\label{eqGradient2us}
\frac{\partial L_{s}}{\partial u_s} = \sum_{t=1}^{k-1} (\mathbf{M}_{t}^T\mathbf{M}_{t})^{-\frac{1}{2}}\mathbf{M}_{t}^T \frac{\partial \mathbf{M}_{t}}{\partial u_s},\nonumber
\\
\frac{\partial L_{s}}{\partial v_s} = \sum_{t=1}^{k-1} (\mathbf{M}_{t}^T\mathbf{M}_{t})^{-\frac{1}{2}}\mathbf{M}_{t}^T \frac{\partial \mathbf{M}_{t}}{\partial v_s},\nonumber
\\
\frac{\partial L_{s}}{\partial u_e} = \sum_{t=1}^{k-1} (\mathbf{M}_{t}^T\mathbf{M}_{t})^{-\frac{1}{2}}\mathbf{M}_{t}^T \frac{\partial \mathbf{M}_{t}}{\partial u_e},\nonumber
\\
\frac{\partial L_{s}}{\partial v_e} = \sum_{t=1}^{k-1} (\mathbf{M}_{t}^T\mathbf{M}_{t})^{-\frac{1}{2}}\mathbf{M}_{t}^T \frac{\partial \mathbf{M}_{t}}{\partial v_e}.
\end{flalign}


These gradients are important since they measure how the sensor length change relates to the change of the design variables (i.e., starting and ending points parameters in $u,v$-domain). Luckily, these gradients can be conveniently calculated by the chain rule according to Eq.~\eqref{eqLinePara} and Eq.~\eqref{eqLineSegment}:

\begin{flalign}
\label{eqGradient2usvsueve}
\frac{\partial \mathbf{M}_{t}}{\partial u_s}  = \mathcolor{red}{\frac{\partial \mathbf{M}_{t}}{\partial u_t} }
\mathcolor{blue}{\frac{\partial u_{t}}{\partial u_s}} + \mathcolor{red}{\frac{\partial \mathbf{M}_{t}}{\partial u_{t-1}} }\mathcolor{blue}{\frac{\partial u_{t-1}}{\partial u_s}}, \nonumber
\\ 
\frac{\partial \mathbf{M}_{t}}{\partial v_s}  = \mathcolor{red}{\frac{\partial \mathbf{M}_{t}}{\partial v_t}} \mathcolor{blue}{\frac{\partial v_{t}}{\partial v_s}} + \mathcolor{red}{\frac{\partial \mathbf{M}_{t}}{\partial v_{t-1}}}\mathcolor{blue}{\frac{\partial v_{t-1}}{\partial v_s}},\nonumber
\\ 
\frac{\partial \mathbf{M}_{t}}{\partial u_e}  = \mathcolor{red}{\frac{\partial \mathbf{M}_{t}}{\partial u_t} }\mathcolor{blue}{\frac{\partial u_{t}}{\partial u_e}} + \mathcolor{red}{\frac{\partial \mathbf{M}_{t}}{\partial u_{t-1}} }\mathcolor{blue}{\frac{\partial u_{t-1}}{\partial u_e}},\nonumber
\\
\frac{\partial \mathbf{M}_{t}}{\partial v_e}  = \mathcolor{red}{\frac{\partial \mathbf{M}_{t}}{\partial v_t} }\mathcolor{blue}{\frac{\partial v_{t}}{\partial v_e}} + \mathcolor{red}{\frac{\partial \mathbf{M}_{t}}{\partial v_{t-1}}} \mathcolor{blue}{\frac{\partial v_{t-1}}{\partial v_e}}.
\end{flalign}

 Note that for the variables in Eq.\eqref{eqGradient2usvsueve} with \textcolor{blue}{blue}, considering that the discretization of the line segment is linear and uniform as shown in Eq.~\eqref{eqLinePara}, they are calculated with the following equation:

\begin{align}
\label{eqPartuv}
\frac{\partial \mathbf{u}_t}{\partial u_s} &= \frac{\partial \mathbf{v}_t}{\partial v_s} = 1 - \frac{t}{K-1}, \nonumber\\
\frac{\partial \mathbf{u}_t}{\partial u_e} &= \frac{\partial \mathbf{v}_t}{\partial v_e} = \frac{t}{K-1}.
\end{align}

For the $\frac{\partial \mathbf{u}_{t-1}}{\partial u_s}$, $\frac{\partial \mathbf{v}_{t-1}}{\partial v_s}$, $\frac{\partial \mathbf{u}_{t-1}}{\partial u_e}$ and $\frac{\partial \mathbf{v}_{t-1}}{\partial v_e}$, the derivation follows the same approach as in Eq.\eqref{eqPartuv}, so the details are omitted for brevity.

 For the variables in Eq.\eqref{eqGradient2usvsueve} with \textcolor{red}{red} are related to the combinations of basis function derivatives. More specifically:

\begin{flalign}
\label{eqPartMtPartUV}
\frac{\partial \mathbf{M}_t}{\partial u_t} = \frac{\partial \mathbf{S}(u_t,v_t)}{\partial u_t} = \sum_{i=1}^{m}\sum_{j=1}^{n} \frac{\partial N_{i,k}(u_{t})}{\partial u_t}N_{j,l}(v_{t}) \mathbf{P}^c_{i,j} \nonumber\\
=\sum_{i=1}^{m}\sum_{j=1}^{n} \frac{\partial N_{i,4}(u_{t})}{\partial u_t}N_{j,4}(v_{t}) \mathbf{P}^c_{i,j} 
\end{flalign}

Note that $k=l=4$ for cublic B-spline surfaces and $\frac{\partial N_{i,4}(u_{t})}{\partial u_t}$ is the B-Spline derivative written as:
\begin{flalign}
\label{eqPartMtPartUV2}
\frac{3}{u_{i+3} - u_{i}}N_{i,3}(u_t) - \frac{3}{u_{i+4} - u_{i+1}}N_{i+1,3}(u_t)
\end{flalign}
Note that $u_{i+3}$, $u_{i}$, $u_{i+4}$ and  $u_{i+1}$ are from the knot vector for B-Spline.
Similarily, $\frac{\partial \mathbf{M}_t}{\partial v_t}$ can be expanded to:

\begin{flalign}
\label{eqPartMtPartUV3}
\frac{\partial \mathbf{M}_t}{\partial v_t} = \frac{\partial \mathbf{S}(u_t,v_t)}{\partial v_t} = \sum_{i=1}^{m}\sum_{j=1}^{n} N_{i,k}(u_{t}) \frac{\partial N_{j,l}(v_{t})}{\partial v_{t}} \mathbf{P}^c_{i,j}\nonumber \\
= \sum_{i=1}^{m}\sum_{j=1}^{n} N_{i,4}(u_{t}) \frac{\partial N_{j,4}(v_{t})}{\partial v_{t}} \mathbf{P}^c_{i,j}
\end{flalign}
where the $\frac{\partial N_{j,4}(v_{t})}{\partial v_{t}}$ is the B-Spline derivative written as:
\begin{flalign}
\label{eqPartMtPartUV4}
\frac{3}{v_{j+3} - v_{j}}N_{j,3}(v_t) - \frac{3}{v_{j+4} - v_{j+1}}N_{j+1,3}(v_t)
\end{flalign}
Also, $v_{j+3}$, $v_{j}$, $v_{j+4}$ and  $v_{j+1}$ are from the knot vector for B-Spline.
For the calculation of $\frac{\partial \mathbf{M}_t}{\partial u_{t-1}}$ and $\frac{\partial \mathbf{M}_t}{\partial v_{t-1}}$, they are similar to Eq.\eqref{eqPartMtPartUV} and Eq.\eqref{eqPartMtPartUV3}. 

It is important to note that Eq.\eqref{eqPartuv}, Eq.\eqref{eqPartMtPartUV}, and Eq.\eqref{eqPartMtPartUV3} are functions of the sensor design variables (see also Eq.\eqref{eqLinePara}). Consequently, the gradient of the sensor length $L_{s}$ depends solely on these design parameters as established by the analytical expressions derived earlier, which makes the backpropagation of sensor length, as shown in Fig.~\ref{figPipeline}, straightforward.

\begin{IEEEbiography}[{\includegraphics[width=1in,height=1.25in,clip,keepaspectratio]{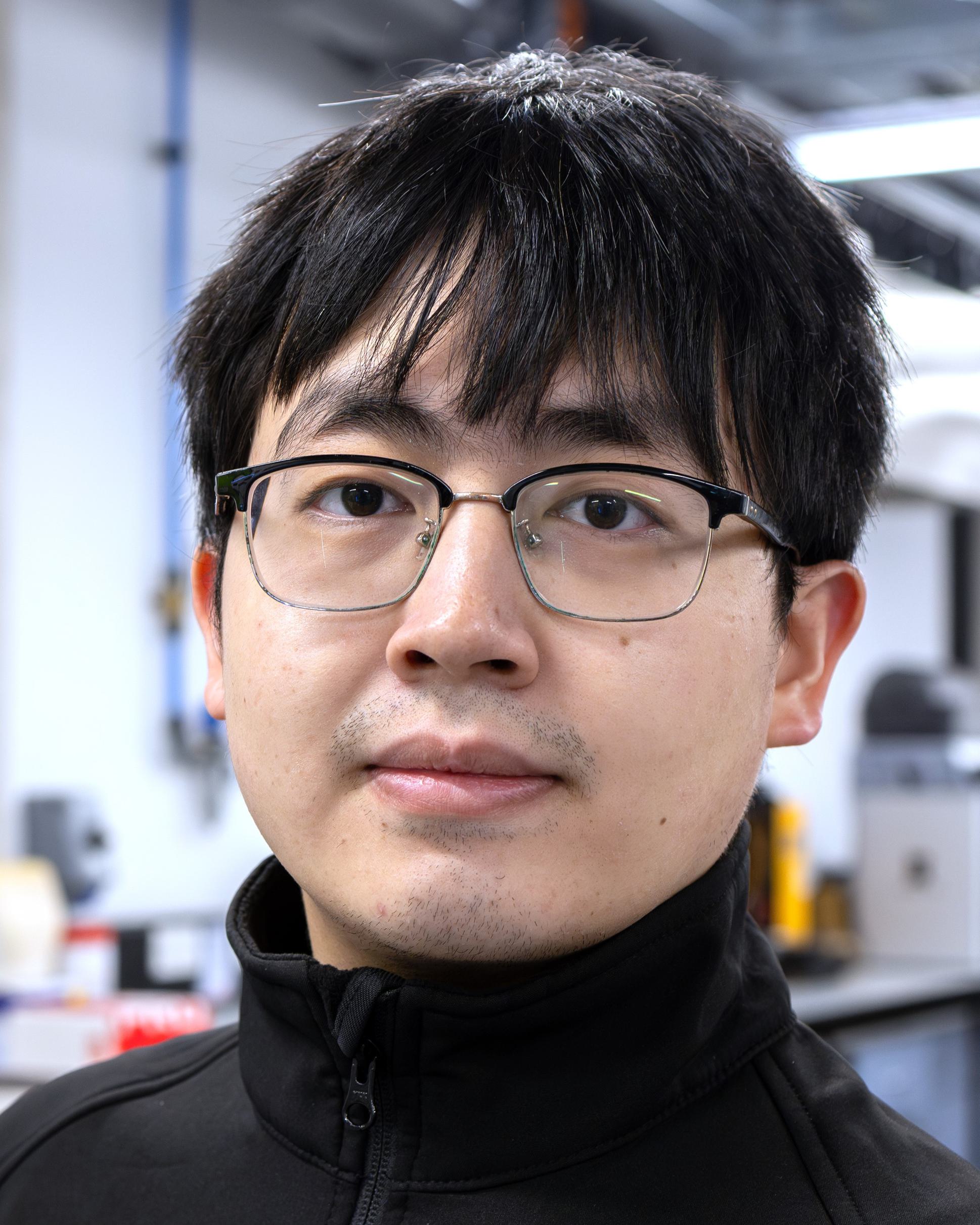}}]{Yingjun Tian} received the B.E. degree in mechanical engineering from the University of Science and Technology of China, Hefei, China, in 2019, and the Ph.D. degree in mechanical engineering from The University of Manchester, Manchester, U.K., in 2024.

He is currently a Postdoctoral Research Associate with the Digital Manufacturing Lab, The University of Manchester. His research interests include computational design, soft robotics, and 3D printing.
\end{IEEEbiography}

\begin{IEEEbiography}[{\includegraphics[width=1in,height=1.25in,clip,keepaspectratio]{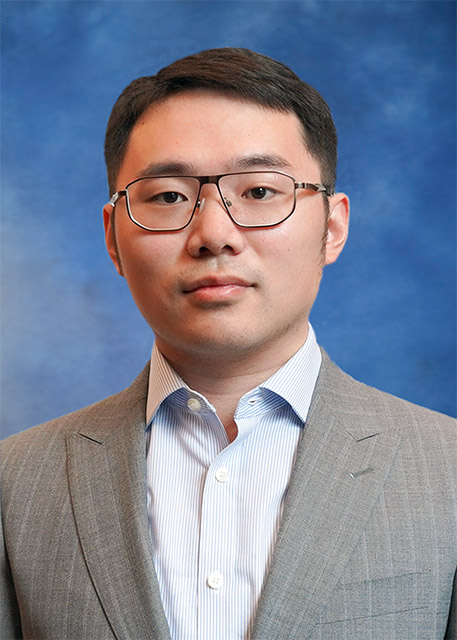}}]{Guoxin Fang} (Member, IEEE) received the B.E. degree in mechanical engineering from the Beijing Institute of Technology, Beijing, China, in 2016, and the Ph.D. degree in advanced manufacturing from Delft University of Technology, Delft, The Netherlands, in 2022. He is currently an Assistant Professor with the Department of Mechanical and Automation Engineering at The Chinese University of Hong Kong, Hong Kong. Prior to this, he was a Research Associate with the Department of Mechanical, Aerospace and Civil Engineering at The University of Manchester, U.K. His research interests include geometric computing, computational design, digital fabrication, and robotics.
\end{IEEEbiography}

\begin{IEEEbiography}[{\includegraphics[width=1in,height=1.25in,clip,keepaspectratio]{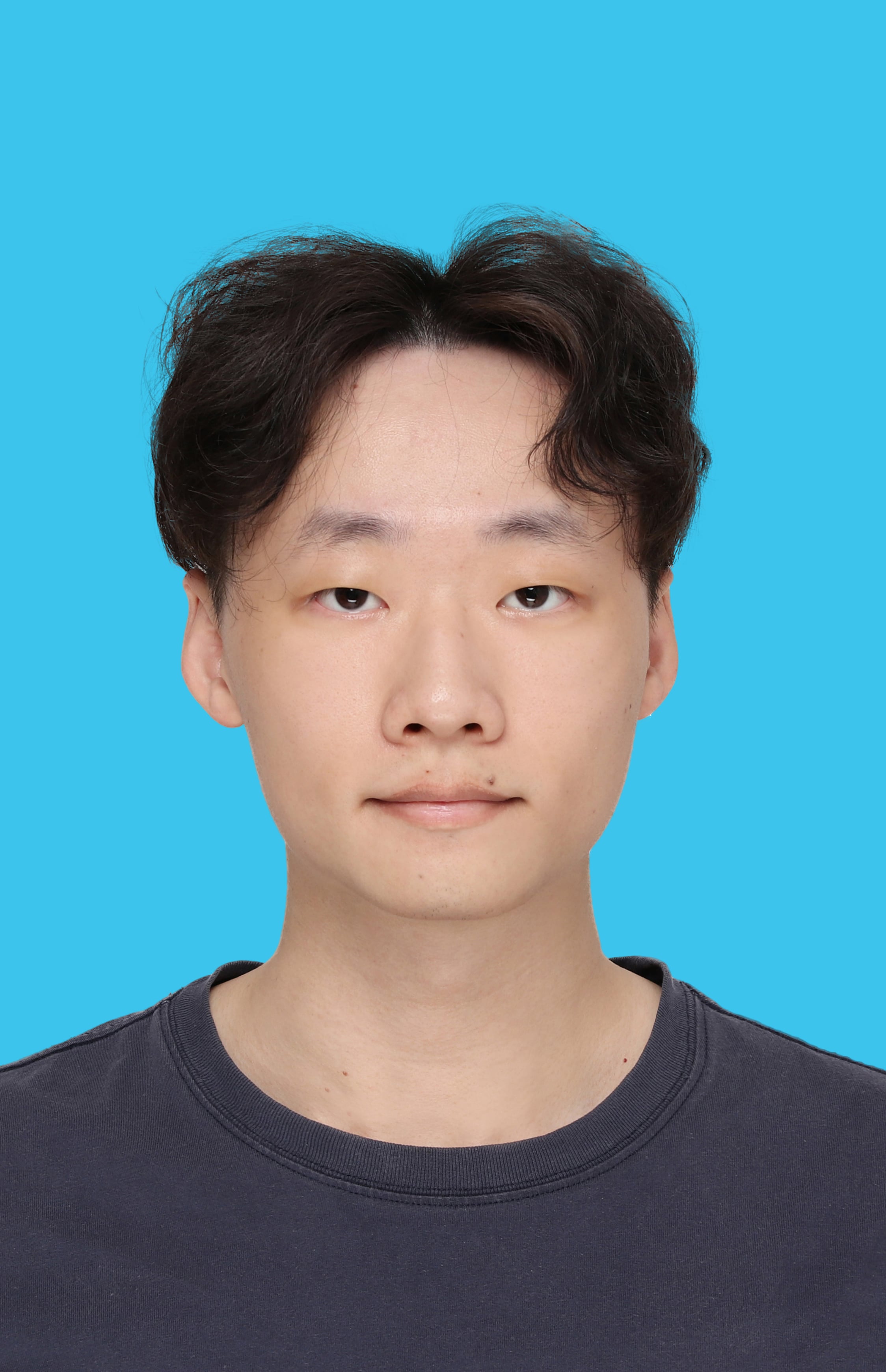}}]
{Aoran Lyu} is currently pursuing a Ph.D. degree in Production and Manufacturing Engineering in the Department of Mechanical and Aerospace Engineering at the University of Manchester. He received his B.S. degree in Mathematics from South China University of Technology in 2021 and his M.E. degree in Computer Technology from the same university in 2024.

His research interests include computer graphics, physics-based simulation, and computational design.
\end{IEEEbiography}

\vspace{-25pt}
\begin{IEEEbiography}[{\includegraphics[width=1in,height=1.15in,clip,keepaspectratio]{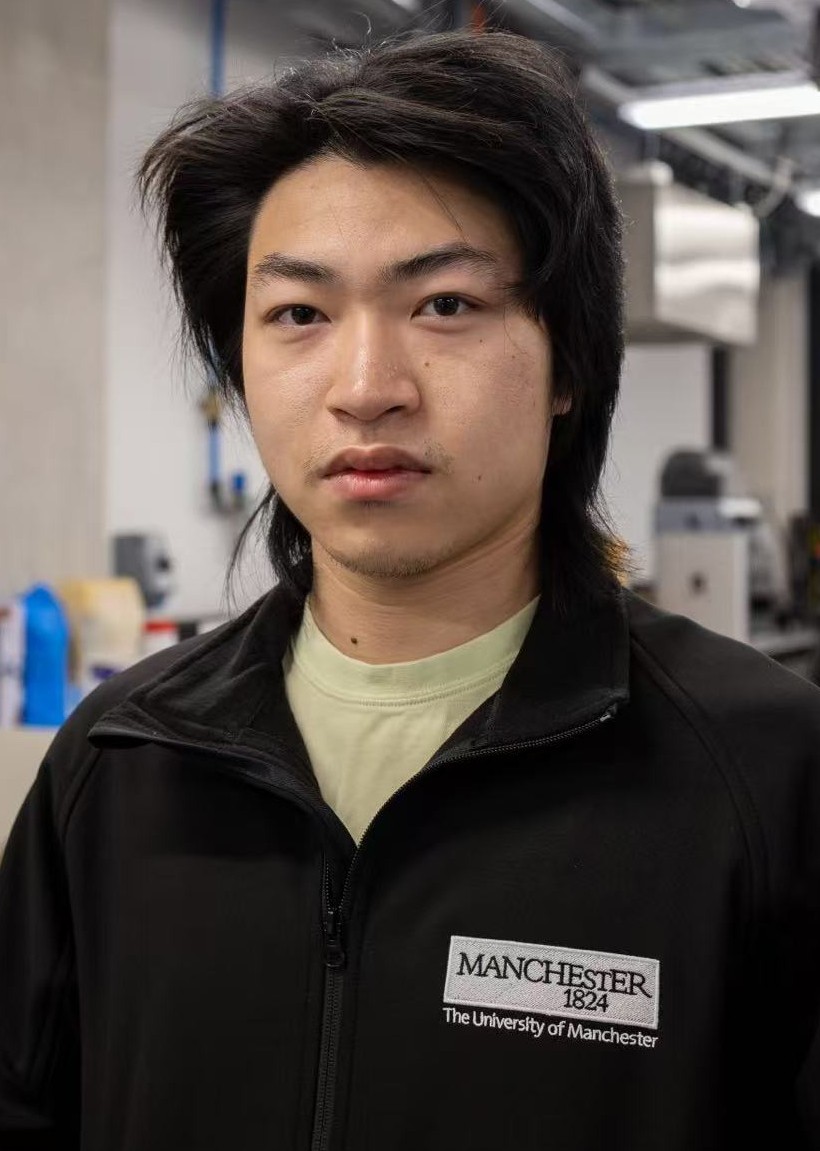}}]
{Xilong Wang} received the B.E. degree in mechanical engineering from The University of Manchester, Manchester, U.K., in 2023, and the MPhil degree in mechanical engineering from the same university in 2026.

His research interests include soft robotics and surface reconstruction.
\end{IEEEbiography}
\vspace{-25pt}

\begin{IEEEbiography}[{\includegraphics[width=1in,height=1.12in,clip,keepaspectratio]{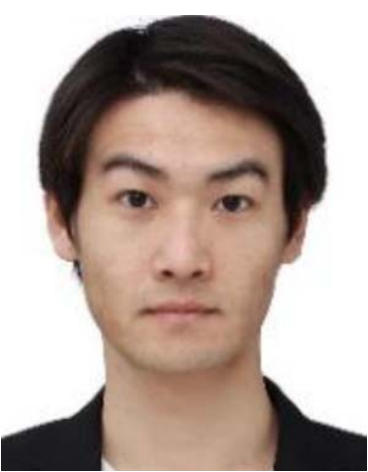}}]
{Zikang Shi} is currently a PhD candidate in the Digital Manufacturing Lab at The University of Manchester. He received his B.Eng. and M.Eng. degrees from the School of Mechanical Engineering at Shanghai Jiao Tong University, China, where he studied from 2017 to 2024. 

His current research interests focus on multi-axis motion planning and robot-assisted manufacturing.
\end{IEEEbiography}
\vspace{-25pt}
\begin{IEEEbiography}[{\includegraphics[width=1in,height=1.18in,clip,keepaspectratio]{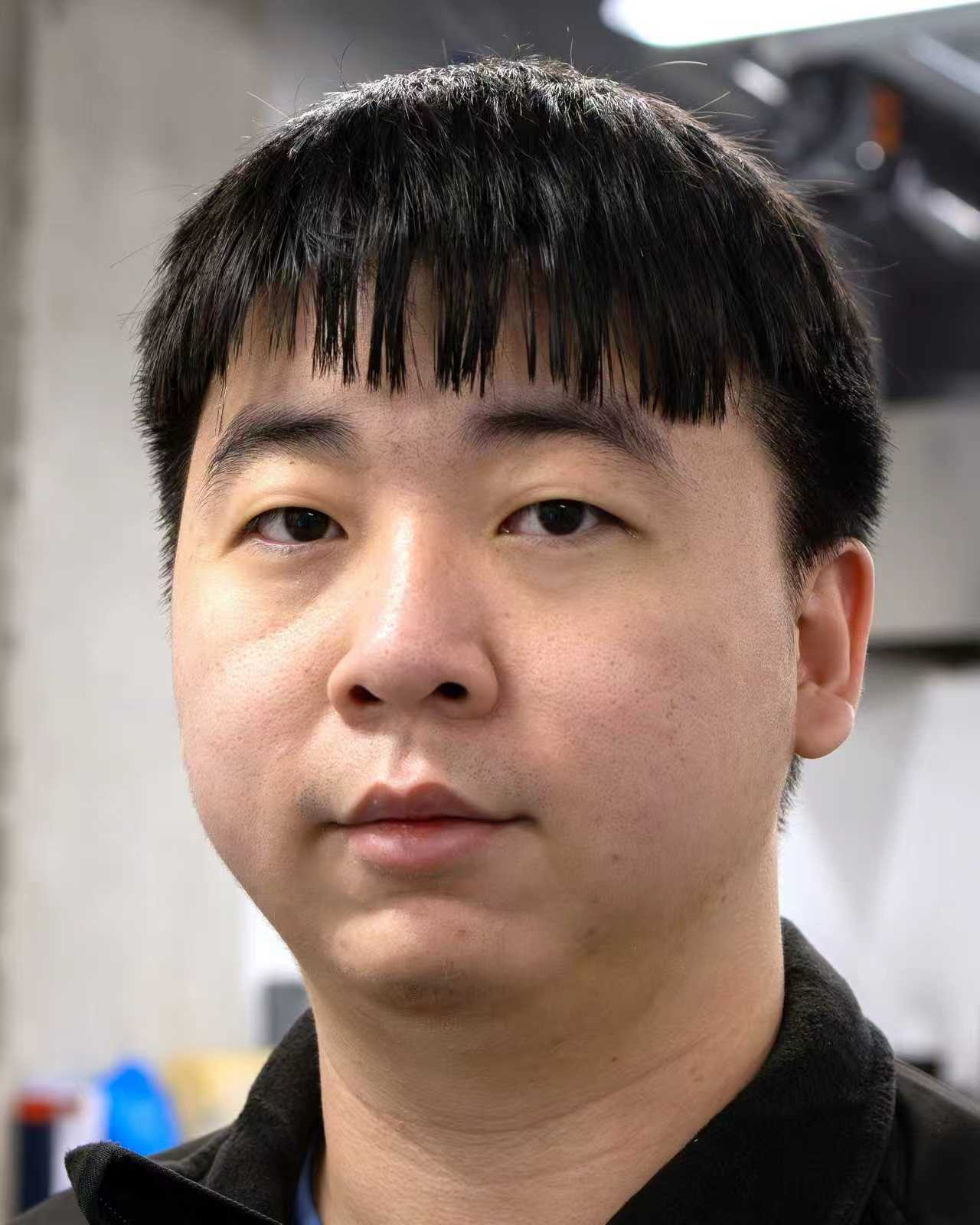}}]
{Yuhu Guo} received the MPhil degree from the University of Manchester, Manchester, U.K. 

His current research interests lie at the intersection of neural graphics, computational design, and robotics, with a focus on developing high-impact methods in learning-based graphics, robotics, and fabrication.
\end{IEEEbiography}

\vspace{-25pt}
\begin{IEEEbiography}[{\includegraphics[width=1.05in,height=1.3in,clip,keepaspectratio]{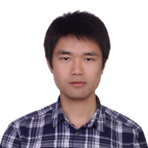}}]
{Weiming Wang} received the BS and PhD degrees from the Dalian University of Technology in 2010 and 2016, respectively.

He was a Postdoctoral Researcher supported by the Marie Skłodowska-Curie LEaDing Fellows Programme at the Delft University of Technology, Delft, The Netherlands. Currently, he is a Postdoctoral Research Associate at The University of Manchester, Manchester, U.K. His research interests are Computational Fabrication and Additive Manufacturing.
\end{IEEEbiography}

\vspace{-25pt}
\begin{IEEEbiography}[{\includegraphics[width=1in,clip,keepaspectratio]{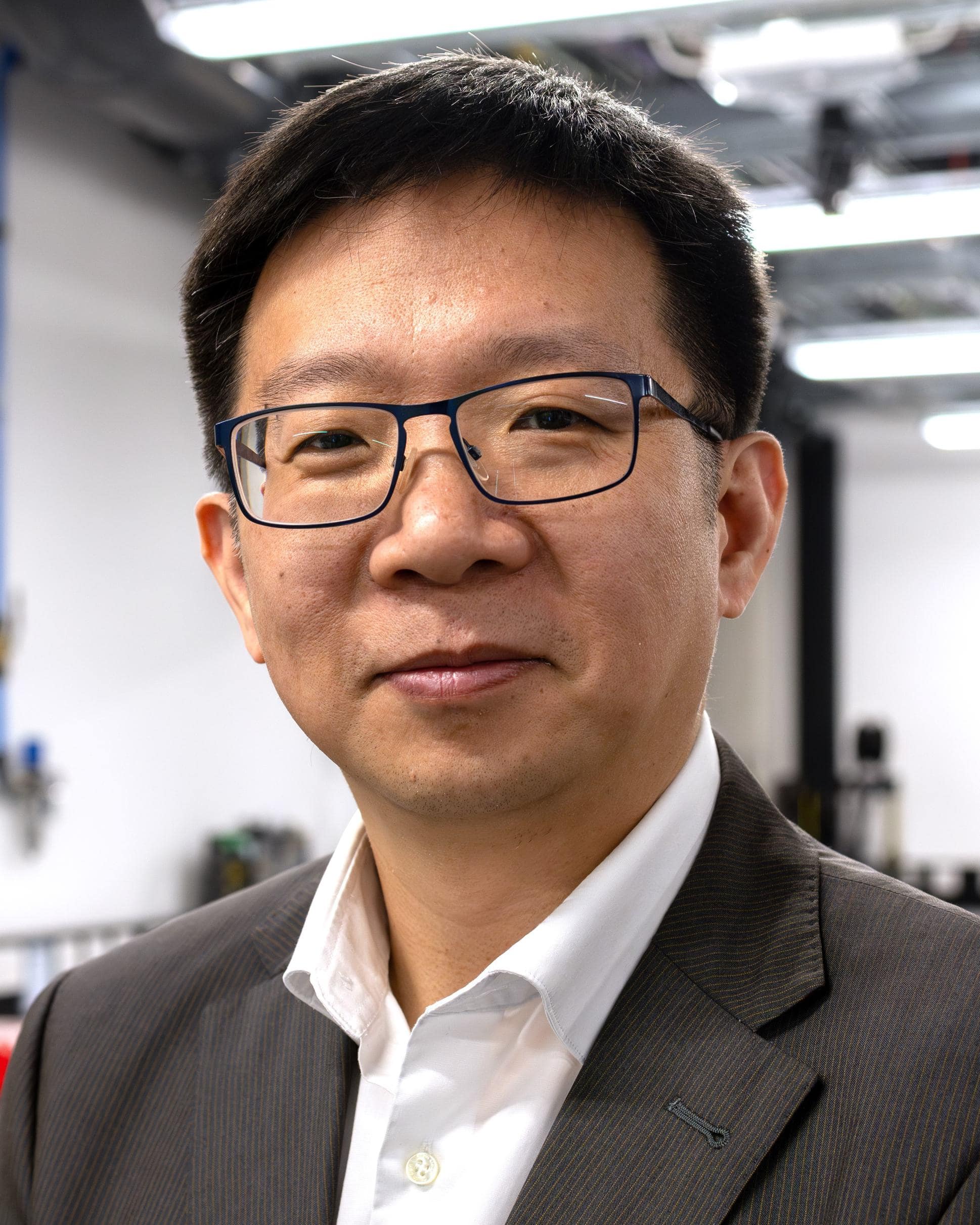}}]{Charlie~C.L.~Wang} is currently Professor and Chair in Smart Manufacturing at the University of Manchester (UoM). Before joining UoM in 2020, he worked as Professor and Chair of Advanced Manufacturing at Delft University of Technology, The Netherlands and as Professor of Mechanical and Automation Engineering at the Chinese University of Hong Kong. He received his Ph.D. degree (2002) in mechanical engineering from Hong Kong University of Science and Technology (HKUST). His research interests include Digital Manufacturing, Computational Design, Additive Manufacturing, Soft Robotics, Geometric Computing, and Computer Graphics. He is Fellow of the American Society of Mechanical Engineers (ASME) and the Solid Modelling Association (SMA).
\end{IEEEbiography}

\end{document}